\documentclass[letterpaper,journal]{IEEEtran}
\IEEEoverridecommandlockouts
\usepackage{booktabs}
\usepackage{cite}
\usepackage{amsmath,amssymb,amsfonts}
\usepackage{algorithmic}
\usepackage{graphicx}
\usepackage{textcomp}
\usepackage{xcolor}
\usepackage{verbatim}
\usepackage{graphicx}
\usepackage{multirow}
\usepackage{pifont}
\usepackage{tabularx}
\usepackage{makecell}
\usepackage[hidelinks]{hyperref} 
\usepackage{array}
\usepackage{soul}
\sethlcolor{yellow!50}
\usepackage[most]{tcolorbox}
\def\BibTeX{{\rm B\kern-.05em{\sc i\kern-.025em b}\kern-.08em
    T\kern-.1667em\lower.7ex\hbox{E}\kern-.125emX}}
\begin{document}

\title{
A Review of Vision-based Vehicle Detection for UAV-based Traffic Monitoring: Experimental Insights and Future Directions
}
\author{Jianlin~Ye and Christos~Kyrkou
\thanks{This work was supported by the European Union’s Horizon Europe research and innovation programme under grant agreement No. 101168067 (GuardAI). Views and opinions expressed are however those of the author(s) only and do not necessarily reflect those of the European Union. Neither the European Union nor the granting authority can be held responsible for them. }
\thanks{The authors are with the KIOS Research and Innovation Centre of Excellence (KIOS CoE), and University of Cyprus, Nicosia, 1678, Cyprus. {\tt\small \{ye.jianlin, kyrkou.christos\}@ucy.ac.cy}}}

\maketitle

\begin{abstract}
\label{sec:abstract}
In Intelligent Transportation System (ITS), unmanned aerial vehicle (UAV)-based surveillance offers an innovative solution to traffic surveillance with wide coverage and real-time data collection capabilities. In comparison to fixed ground-based infrastructure, UAVs are able to respond to dynamic traffic but present challenges such as vehicle detection at varying altitudes, compensation for motion-induced image variations and efficient processing of high-resolution images. Deep learning has been largely beneficial on improving the detection accuracy; however, for practical deployment, a critical assessment of the accuracy, latency, and harmonization with current transportation systems needs to be carefully considered. This survey reviews recent advancements in the UAV-based traffic monitoring, with a primary focus being deep neural network models for traffic analytics in various urban settings. Three main challenges identified in the literature are ensuring compatibility with traffic control systems, achieving real-time processing to optimize traffic flow, and maintaining robust detection in different environmental conditions. Existing solutions often lack comprehensive frameworks for utilizing UAV captured data to respond to incidents and manage traffic effectively. Future research should focus on optimal detection models, edge processing, and adaptive control integration to improve the responsiveness of urban traffic management. 
\end{abstract}

\section{Introduction}
\label{introduction}

\IEEEPARstart{I}{ntelligent} Transportation Systems (ITS) have played a pivotal role in enhancing urban mobility and road safety. Over one million annual road deaths globally~\cite{zhang2011data} underscores the critical need for effective traffic monitoring and rapid-response systems~\cite{kuznietsov2024explainable}. As a key emerging technology, Unmanned Aerial Vehicles (UAVs) provide comprehensive, real-time aerial surveillance, enabling dynamic ground tracking even across complex road networks~\cite{savkin2021navigation}. This paper focuses on a specific application of this technology: UAV-enabled traffic monitoring within the ITS framework, and examines the state-of-the-art vehicle detection techniques fundamental to these systems. 

\begin{table*}[t]
\centering
\caption{Comparison of focus areas between existing surveys and our work. A checkmark (\checkmark) indicates a primary focus, while a dash (-) indicates areas outside the main scope of the respective survey.}
\resizebox{\textwidth}{!}{
\begin{tabular*}{\textwidth}{@{\extracolsep{\fill}} l c c c c c c l}
\toprule
\textbf{Reference} & \multicolumn{6}{c}{\textbf{Focus Areas}} & \textbf{Primary Contribution / Summary} \\
\cmidrule{2-7}
& \textbf{Monitoring} 
& \textbf{Detection} 
& \textbf{Tracking} 
& \textbf{\begin{tabular}[c]{@{}c@{}}Data\\Fusion\end{tabular}} 
& \textbf{\begin{tabular}[c]{@{}c@{}}Deployment\\Constraints\end{tabular}} 
& \textbf{\begin{tabular}[c]{@{}c@{}}Real-time\\Analysis\end{tabular}} & \\
\midrule
\addlinespace
Menouar et al. \cite{menouar2017uav} & \checkmark & - & - & - & \checkmark & - & UAV-ITS macro-architecture \\
Outay et al. \cite{outay2020applications} & \checkmark & - & - & - & \checkmark & - & Road infrastructure monitoring \\
Kumar et al. \cite{kumar2021study} & - & \checkmark & - & - & - & \checkmark & Detection algorithms \\
Gupta et al. \cite{gupta2021advances} & \checkmark & - & - & - & \checkmark & - & UAVs in future transportation \\
Bouguettaya et al. \cite{bouguettaya2021vehicle} & - & \checkmark & - & - & \checkmark & \checkmark & DL for aerial detection \\
Telikani et al. \cite{telikani2024machine} & - & \checkmark & - & - & \checkmark & \checkmark & ML for perception layer \\
Xiao et al. \cite{xiao2025uav} & \checkmark & \checkmark & - & - & - & \checkmark & Oriented bounding boxes \\
Bisio et al. \cite{bisio2022systematic} & \checkmark & \checkmark & \checkmark & - & \checkmark & \checkmark & End-to-end monitoring systems \\
Telikani et al. \cite{telikani2025unmanned} & \checkmark & \checkmark & \checkmark & - & \checkmark & \checkmark & Advances in ITS monitoring \\
\midrule
\textbf{Ours (This Survey)} & \checkmark & \checkmark & \checkmark & \checkmark & \checkmark & \checkmark & Algorithms \& deployment synthesis \\
\bottomrule
\end{tabular*}
}
\label{tab:uav-its-survey}
\end{table*}

Traffic monitoring by use of UAVs has specific benefits over ground monitoring techniques. The high altitude locations guarantee full coverage of cities, enhancement of visibility in crowded areas and quality to provide high-resolution images to track traffic movement comprehensively and to overcome occlusion issue experienced in ground-level sensors~\cite{li2023developing}. Together with developments in parallel computing, including multi-core processing units and graphics processing units (GPUs) the systems are currently capable of having complex deep learning models deployed to further boost vehicle detection accuracy in real time. Current research by Sun et al.~\cite{sun2024efficient} shows that multi-UAV coordination can achieve an even better coverage and data-processing efficiency in large-scale traffic situations. However, the successful integration of these deep learning outputs into the broader ITS is constrained by latency and processing requirements. Realizing such real-time cooperative systems requires both optimizing the computational complexity of the CNN models and addressing the routing challenges inherent in integrating UAVs with dynamic urban Vehicular Ad-hoc Networks (VANETs). As demonstrated by recent network frameworks, utilizing cooperating UAVs for on-demand data transmission is necessary to ensure that vision-based inferences are delivered to ground infrastructure with minimal delay~\cite{oubbati2018demand}.

The recent advancements in deep learning, particularly Convolutional Neural Networks (CNNs) have significantly improved vehicle detection in terms of accuracy and robustness. Single shot object detection algorithms like You Only Look Once (YOLO) detection framework have shown to be highly effective for analyzing UAV-captured imagery by providing simultaneous detection and tracking capabilities~\cite{redmon2016you}. However, one current research challenge is to adapt these models to the specific UAV-based scenarios, including perspective changes, occlusion, and dynamic environments. These challenges are leverage using emerging solutions that leverage multi-UAV collaboration~\cite{sun2024efficient}.

This survey reviews the literature on vehicle detection technologies for UAV-based traffic monitoring in ITS. Focusing primarily on studies published after 2022 and building upon pioneering work~\cite{menouar2017uav, outay2020applications, sivaraman2011combining, kumar2021study, gupta2021advances, bouguettaya2021vehicle, telikani2024machine, xiao2025uav, bisio2022systematic, telikani2025unmanned}, we analyze the evolution from traditional vision-based approaches to current deep learning strategies. By bridging theoretical advances and practical implementations, this survey aims to provide valuable insights into the development of more efficient and responsive UAV-based traffic monitoring systems within ITS. Our contributions are as follows:

\begin{itemize}
    \item We outline the architectural design elements of the existing detection models and provide an in-depth investigation of various YOLO algorithms and variants tailored for UAV applications.
    \item We assess the recent benchmark datasets and performance evaluation with comparison experiments on a few distinct datasets that simulate real-world traffic scenarios.
    \item Finally, we outline future research directions aimed at promoting further development and addressing key challenges in UAV-based ITS applications.
\end{itemize}

The rest of the survey is organized as follows. The related works in the field are reviewed in Section~\ref{sec:related_work}. Section~\ref{sec:challenges} discusses the open problems and potential challenge of vehicle detection, tracing the evolution from traditional to deep learning methods. In Section~\ref{sec:architectural_components}, we examine the architectural elements of modern detection models, while Section~\ref{sec:yolo_variants} studies variations of the YOLO algorithm specifically tailed in UAV-based vehicle detection. Section~\ref{sec:benchmark_datasets} provides benchmark datasets, evaluation metrics, and Section~\ref{sec:comparative_experiments} performs comparative experiments on selected datasets. Section~\ref{sec:future_directions} provides future research directions in the ITS domain, and Section~\ref{sec:conclusion} concludes the paper.

\begin{figure*}[htbp]
    \centering
    \includegraphics[width=0.7\textwidth]{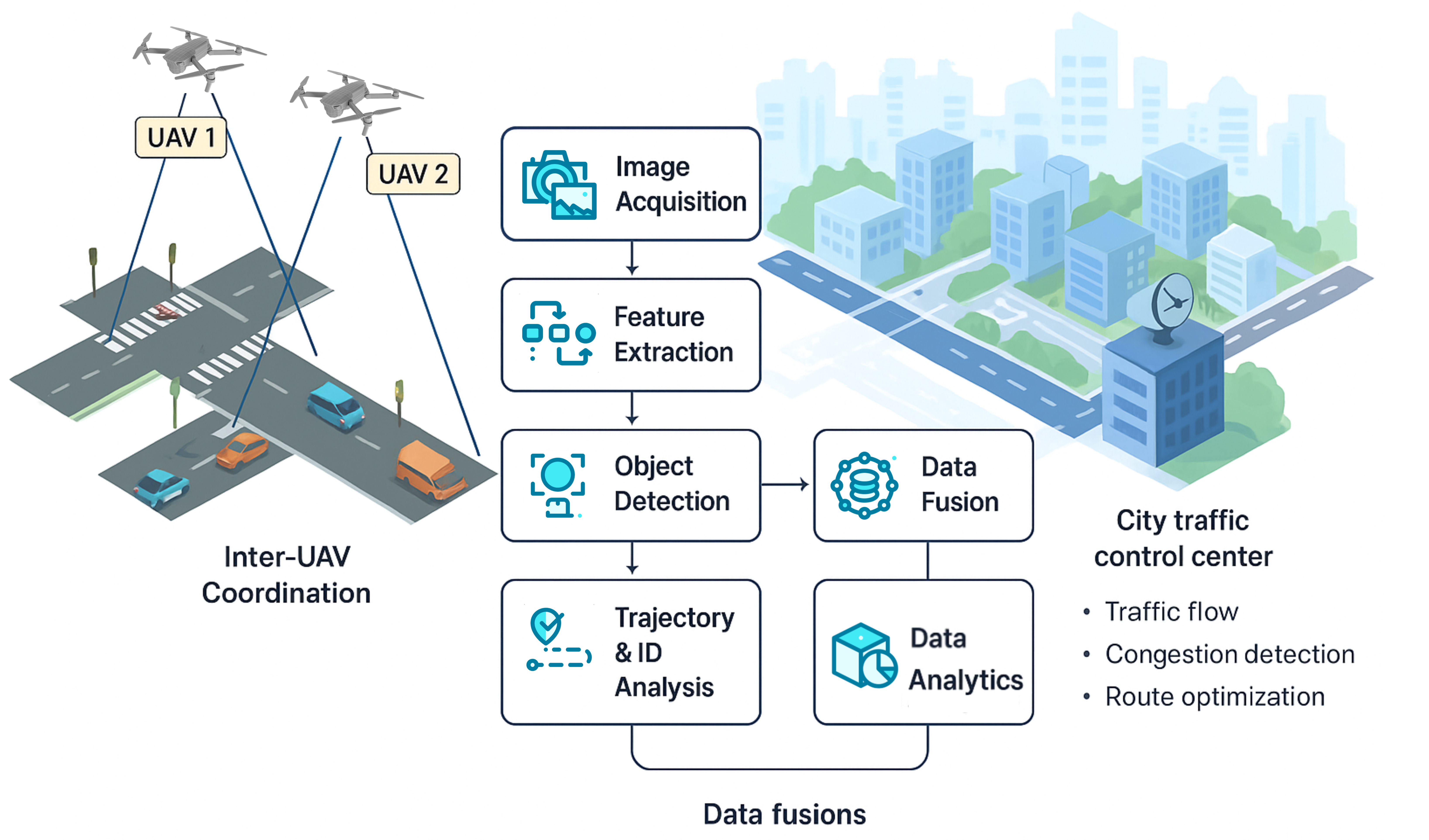}
    \caption{Overall workflow of the UAV-based intelligent traffic monitoring system. Multiple UAV nodes acquire real-time images of road intersections, perform feature extraction, vehicle detection, and tracking, followed by trajectory and ID analysis. Results from different UAVs are fused through inter-UAV coordination, and the fused data is analyzed at the city traffic control center for applications such as traffic flow estimation, congestion detection, and route optimization.}
    \label{fig:workflow}
\end{figure*}
\section{Related Work}
\label{sec:related_work}
This section provides an overview of existing survey papers and related academic works in the domain of UAV-based traffic monitoring. As summarized in Table~\ref{tab:uav-its-survey}, while prior surveys have addressed specific aspects of UAV-aided ITS, our work presents a comprehensive synthesis that integrates these diverse topics.

\subsection{System-Level UAV Applications in ITS}
The current literature on UAV applications in ITS can be broadly categorized into two main research streams. The first stream focuses on the system-level application of UAVs within transportation infrastructure. Menouar et al.~\cite{menouar2017uav} and Outay et al.~\cite{outay2020applications} established foundational frameworks for UAV deployment in urban mobility. These foundational works primarily concentrated on monitoring applications, Telikani et al.~\cite{telikani2025unmanned} extended the discussion to cover transmission and detection paradigms.

This stream is further depicted by a number of studies. Butilua et al.~\cite{butilua2022urban} reviewed UAV applications for traffic monitoring. Srivastava et al.~\cite{srivastava2021survey} analyzed important parameters concerning the small object detection. The analyses of UAV remote sensing imagery and deep learning (DL) techniques were conducted by Osco et al.~\cite{osco2021review}, whereas Alzahrani et al.~\cite{alzahrani2020uav} reviewed UAV-assisted systems currently available. Kanistras et al.~\cite{kanistras2013survey} addressed the monitoring of traffic by UAV, and Outay et al.~\cite{outay2020applications} discussed vision processing methods. Park et al.~\cite{park2020application} studied DL in real-time applications UAV-based traffic analysis, whereas Zhang et al.~\cite{zhang2022analysis} focused on approaching the problem of target occlusion using correlation filtering and tracking algorithms.

\subsection{Deep Learning–Based Vehicle Detection}
In parallel, a second research stream addresses algorithmic advancements in vehicle detection. Kumar et al.~\cite{kumar2021study} and Bouguettaya et al.~\cite{bouguettaya2021vehicle} focused on the feature extraction technique and DL architecture. Liu et al.~\cite{liu2019performance} compared four vehicle detection models: Faster R-CNN, YOLOv3, a Support Vector Machine with Histogram of Oriented Gradients features and the Visual Background Extractor algorithm. Their results showed that YOLOv3 and Faster R-CNN performed the best but the latter required considerably more hardware resources even with greater recall and precision. The perspectives of DL on the UAV-based traffic monitoring are highlighted in other studies. Bisio et al.~\cite{bisio2022systematic} reviewed drone-based traffic monitoring systems, covered vehicle detection, tracking, and counting in smart cities, and mentioned some challenges, such as varying object scales, viewing angles, and occlusions. Iftikhar et al. ~\cite{iftikhar2023target} reviewed DL-based approaches for analysis traffic congestion using aerial imagery and videos. Telikani et al. ~\cite{telikani2024machine} discussed the role of UAVs with machine learning (ML) and DL in increasing the monitoring of traffic, responding in cases of emergencies, and inspecting infrastructure. They tested models such as CNN, R-CNN, Faster R-CNN and YOLO on different continuously collected aerial vehicle and pedestrian detection datasets. Extensive DL perception for UAV in-depth analyses was also done by Xiao et al. ~\cite{xiao2025uav}.

\subsection{Recent Comprehensive Reviews}
A number of recent surveys have further developed perception and traffic monitoring using the UAV. Tang et al.~\cite{tang2023survey} conducted a deep learning–focused review on UAV object detection, discuss the challenges of small size of objects, rotation and scale variation, and presents a list of the current solutions to these problems including transformer-based and attention-based models. Wu et al.~\cite{wu2021deep} conducted a survey of methods in detection and tracking from UAV videos, categorizing hundreds of methods and summarizing landmark datasets and thematic issues. Li et al.~\cite{li2024comparison} compared DL architectures and pointed out trade-offs in detection accuracy and real-time performance. Nikouei et al.~\cite{nikouei2025small} described methods of small-object detection, including multi-scale fusion, super-resolution, attention, or transformer networks that address vehicle-scale object monitoring challenges in UAV monitoring. 

\subsection{Research Gaps}
Along with the advancements in UAV-based monitoring of traffic, some key challenges have been under-explored. Research that already exists on detection accuracy typically leaves out considerations related to system-level integration, especially, the real-time processing requirements and parameters extraction of traffic. The summary Table~\ref{tab:uav-its-survey} shows three existing gaps that need to be bridged further in a unified manner: 
\begin{enumerate}
    \item the integration of UAV mobility restrictions into detection-to-analysis pipelines,
    \item further adaptation of tracking algorithms to oblique aerial views, 
    \item the lack of overall performance evaluation frameworks with requirements of the operational environment, like the computational constraints and the variance of the environment.
\end{enumerate}

In this setting, two domain specific gaps are more urgent concerning multi-UAV deployments. To begin with, there are inadequate methodologies on multi-UAV fusion detection outcomes which end up yielding redundant detections, incomplete spatial coverage, and loss of accuracy in overlapping areas. Second, cross-UAV vehicle tracking is an open problem with regularly occurring challenges of regulating the vehicle identities under varying camera viewpoints, varying illumination, and partial occlusions. Such circumstances have to be resolved in order to provide viable and scalable real-time multi-UAV traffic monitoring systems that can provide dependable intelligence in large scale operational areas.
\section{Architectural Components of Detection Models}
\label{sec:architectural_components}
\begin{figure*}[htbp]
    \centering
    \includegraphics[width=0.75\linewidth]{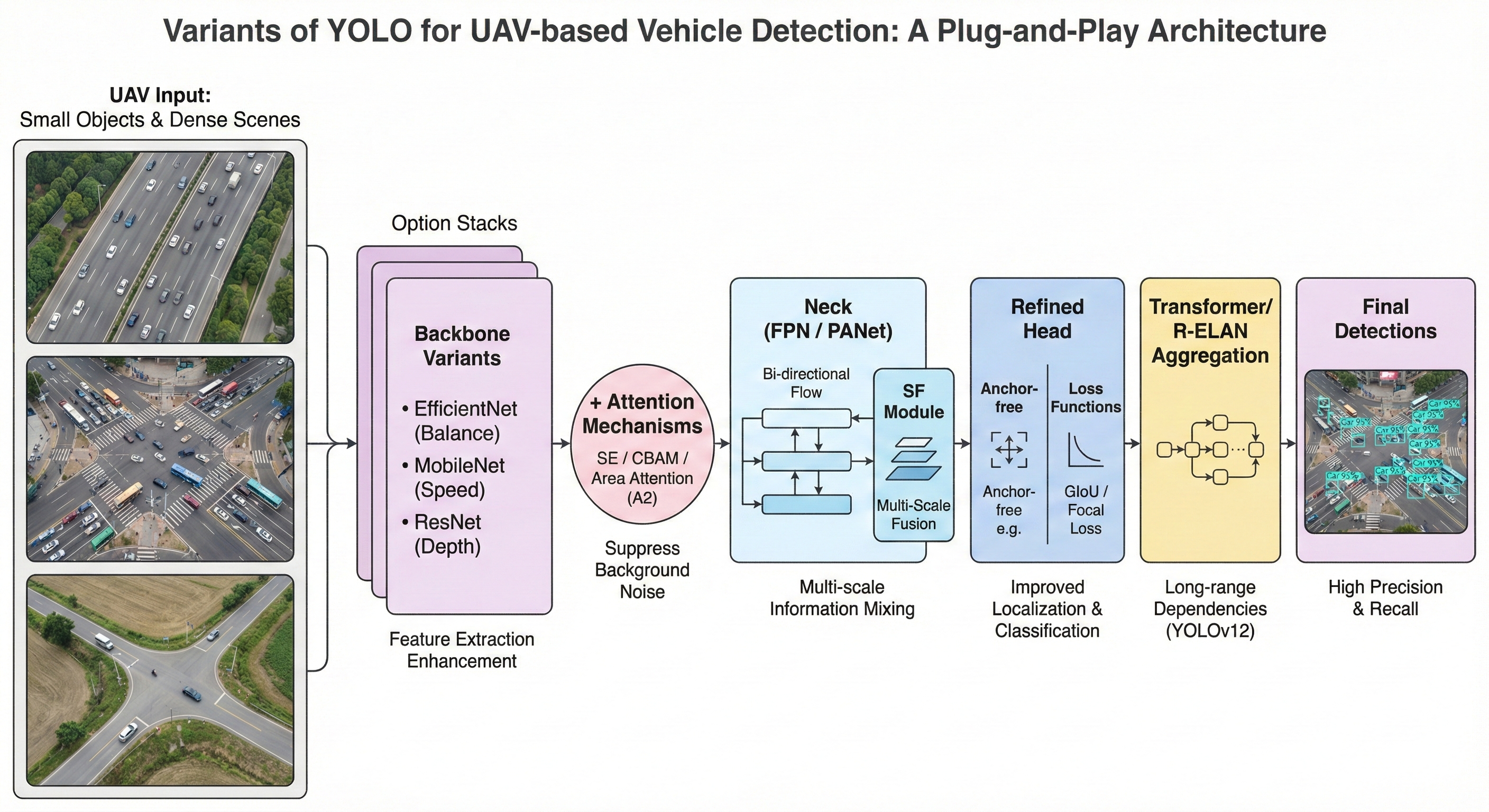}
    \caption{A generic architectural workflow for vision-based UAV traffic monitoring. The process begins with preprocessing to handle high-resolution inputs. The Backbone extracts features at multiple scales, which are fused in the Neck module to address the challenge of small object detection (scale diversity). The Head decouples classification and localization tasks. Finally, to ensure temporal consistency required for traffic monitoring, a Data Association / Tracking Module assigns unique IDs to detected vehicles, converting static detections into dynamic trajectories.}
    \label{fig:Architecture}
\end{figure*}

UAV-based vehicle detection faces unique challenges, including small object sizes, complex environmental backgrounds, and real-time processing requirements. To provide a structural foundation for understanding the solutions discussed in this section, Fig.~\ref{fig:Architecture} illustrates a generic reference architecture for vision-based UAV traffic monitoring. This diagram maps the end-to-end data flow from input preprocessing and feature extraction to the final trajectory generation.

\subsection{Backbone Networks}
The backbone networks form the foundation of the object detection networks, since they are major feature extractors, which convert input images to hierarchical feature maps. Such maps are further on processed by the other components (e.g., neck and head modules) to locate the object and its classification.  In UAV-based vehicle detection, the backbone should support a good feature representation and computational efficiency since both have a direct influence on the accuracy of the detection and the real-time performance. Various backbone architectures are common in use. For instance, ResNet~\cite{he2016deep}  is famous for its residual deep design on the reduction of the gradient degradation and improvement of robust feature learning. VGG~\cite{simonyan2014very} presents a simple design and an effective feature-hierarchy extraction. Instead, MobileNet~\cite{howard2017mobilenets} uses depthwise separable convolutions to significantly reduce computational cost, making it well-suited for resource-limited UAV platforms.

To achieve the balance of efficiency and precision in UAV applications, MobileNet is often combined with task-specific frameworks such as EdgeNet~\cite{plastiras2019edgenet}. EdgeNet dynamically filters low-priority regions in high-resolution UAV imagery, achieving up to a 100 times faster data reduction without affecting detection accuracy. This architecture is capable of making real-time inference within less than 4W of power consumption on embedded processors, a feature needed because of the limited power requirements of UAVs. For UAV deployment, DroNet~\cite{kyrkou2018dronet} introduces a lightweight CNN backbone optimized for real-time inference on embedded processors, achieving a balance between accuracy and efficiency. The combination of the lightweight backbones and optimizations specific to different requirements enables researchers to cope with the computational efficiency and detection reliability issues of aerial surveillance efficiently.

\subsection{Neck Components}
The neck part of a detection model boosts feature fusion between the backbones of various levels. This is crucial when applied in detecting vehicles where objects vary in aspect ratio, orientation, and size. The two very popular neck architectures are Feature Pyramid Networks (FPN) and Path Aggregation Networks (PANet)~\cite{lin2017feature, liu2018path}. FPN enhances the detection by building a multi-scale feature pyramid so that detections can be more effective across scales. PANet extends this, improving crosstalk between pyramid layers, improving the detection of small and occluded vehicles~\cite{liu2018path}. These neck architectures are highly effective and contribute significantly to the refinement of the vehicle detection model accuracies as well as model robustness especially in complex scenes with cluttered background or a varying lighting~\cite{tan2020efficientdet}.

\subsection{Head Modules}
The head module performs object classification and localization using feature maps from the backbone and neck components. Detection heads are classified into single-stage types, such as YOLO~\cite{redmon2016you} and SSD~\cite{liu2016ssd}, and two-stage types, such as Faster R-CNN~\cite{ren2016faster}. Single-stage heads produce bounding boxes and class probabilities in a single pass giving faster, real-time performance. Nevertheless, they could compromise a little in terms of accuracy compared to two-stage heads which first propose regions and then optimize them. 

A major advancement in head architecture is the use of decoupled heads, as in YOLO-X~\cite{ge2021yolox}. Conventional versions of YOLO use coupled heads, where classification and regression share feature representations. YOLO-X decomposes these tasks into separate branches, enabling each to optimize for its specific goal—either classification or bounding box regression. This design enhances performance through minimizing conflicts between the two objectives, especially the small object detection and the minimizing the classification-regression contradiction that is prevalent in vehicle detection.  

Recent works have developed hybrid head architecture that utilizes the speed of single-stage detection and accuracy of two-stage detection giving us a potential direction to future vehicle detection development~\cite{tian2019fcos}. Combined with decoupled head demonstrate the field's progression towards detection architectures that are both accurate and computationally efficient.

\subsection{Multi-Scale Detection Strategies}
Vehicle monitoring using UAVs requires multi-scale detection since the scale of the object varies considerably under the influence of a change in the altitude of the device and the perspective. Core neck models FPN~\cite{lin2017feature}, PANet~\cite{liu2018path} incorporate scale diversity through the combination of hierarchical features. Early work by Plastiras et al.~\cite{plastiras2018efficient} proposed tile-based processing of UAV, where selective attention features were applied, to focus on small objects found in high-resolution image patches. The approach enhanced the accuracy of detection without sacrificing computational efficiency, paving the way for modern slice-aided frameworks.

Additionally, to enhance the process of small object detection, networks like Slicing Aided Hyper Inference (SAHI)~\cite{akyon2022slicing} are designed externally to the architecture to enhance existing designs. The slice-aided inference and fine-tuning pipeline implemented in SAHI processes high-resolution UAV captured image data in patches of overlapping areas, increasing details of small vehicles that detectors would not pick up. For example, SAHI can increase detection AP by up to 14.5\% on aerial datasets such as VisDrone when applied to models like YOLOv5 or FCOS \cite{akyon2022slicing}. In contrast to FPN and PANet, which alter internal feature fusion, SAHI operates as pre-processing and post-processing wrapper, and is thus detector-agnostic and highly adaptable to UAV-specific challenges.

Recent architectural developments, including PC-YOLO11s~\cite{wang2025pc}, refine the hierarchies of backbones with regard to detecting small objects. This model also includes a coordinate spatial attention which highlights both spatial and positional indicators, hence minimizing the background interference. Its broad applicability is tested in UAV and agricultural data demonstrating its effectiveness in the practical application of small object detection. Similarly, LUD-YOLO~\cite{fan2025lud} introduces a multi- scale feature fusion strategy that combines up-sampling with a progressive Feature Pyramid Network, improving small object detection while maintaining a lightweight design suitable for UAV deployment. These studies illustrate the evolution of multi-scale strategies, from architectural modifications to hybrid frameworks, ensuring robust vehicle detection detection abilities under UAV operating in a variety of conditions.

\subsection{Transformers-based vs. CNN-based Models}
The architectural differences between Transformer-based models and traditional CNNs represent a significant shift in detection model design. Transformers were originally built to do natural language processing but have since been modified to computer vision where they provide some benefits over CNNs, especially in setting long-range dependencies and global context~\cite{carion2020end}.   

In contrast to CNNs, which perform hierarchical feature extraction through Convolutional layers, Transformers apply self-attention to learn the relative importance of input components and to capture relations between pixels in remote locations better~\cite{zhu2020deformable, zong2023detrs, zhao2024detrs, minderer2022simple, zhang2022dino}. It is particularly useful in vehicle detection activities involving complicated scenes when context plays a primary role in object detecting.

However, their great computational cost restricts its implementation in real-time or resource-limited areas like UAV-based vehicle detecting~\cite{khan2022transformers}. Despite these constraints, Transformers are quite promising, and current research efforts are focused on the optimization of their efficiency without compromising good performance. New trends indicate that the hybrid model that combines CNN and Transformer models may provide high accuracy and efficiency in vehicle detection models.

To bridge the gap between accuracy and latency, specific architectures like the Swin Transformer and RT-DETR have emerged as strong contenders. The Swin Transformer serves as a hierarchical backbone that computes self-attention within non-overlapping shifted windows. This design significantly reduces computational complexity while retaining global context modeling, offering a critical advantage for distinguishing vehicles in dense urban clutter where CNNs might struggle with occlusion \cite{liu2021swin}. Furthermore, RT-DETR addresses the slow convergence of original DETR models by introducing an efficient hybrid encoder, achieving accuracy comparable to or exceeding YOLO variants on standard benchmarks~\cite{zhao2024detrs}. Most recently, RF-DETR has been introduced by Roboflow as a state-of-the-art evolution, further optimizing the transformer architecture to reduce latency while maintaining the semantic understanding required for complex scenarios~\cite{robinson2025rf}.

Despite these advancements, a critical trade-off remains for UAV deployment. While Transformer-based models offer superior global receptive fields, they generally impose a higher memory footprint. On resource-constrained edge devices (e.g., NVIDIA Jetson Orin), highly optimized CNN architectures like YOLO continue to offer a superior frames-per-second (FPS) to energy consumption ratio~\cite{li2022efficientformer, bianco2018benchmark}. Consequently, while Transformers represent the future of high-accuracy monitoring, CNNs remain the pragmatic choice for onboard, real-time aerial processing where battery life is the limiting factor.


\subsection{Integration of Detection and Tracking} 
While YOLO-based models excel at spatial localization, comprehensive traffic monitoring inherently requires temporal consistency to estimate flow and trajectories. The Tracking-by-Detection (TBD) paradigm addresses this by linking per-frame bounding boxes to existing tracklets via a dedicated association module, forming the standard architecture for UAV surveillance.

\subsubsection{Motion and Appearance-based Tracking (DeepSORT \& StrongSORT)} DeepSORT~\cite{wojke2017simple} serves as a foundational TBD algorithm, integrating Kalman filtering for motion prediction with Deep Appearance Descriptors (Re-ID) to mitigate identity switches during occlusion. However, rapid UAV ego-motion and abrupt viewpoint changes often degrade standard Kalman filter performance. To address this, StrongSORT~\cite{du2023strongsort} advances the methodology by incorporating Camera Motion Compensation (CMC) for spatial alignment and an improved feature extractor (BoT-SORT), offering superior robustness against dynamic aerial flight characteristics.

\subsubsection{Association-based Tracking (ByteTrack)} Traditional trackers often discard low-confidence detections to reduce false positives, leading to track fragmentation during occlusion or motion blur—common issues in aerial imagery. ByteTrack~\cite{zhang2022bytetrack} resolves this via the BYTE association strategy, which prioritizes matching high-confidence detections before recovering low-confidence boxes to reconstruct trajectories. By capitalizing on the high localization accuracy of modern detectors rather than relying solely on computationally expensive Re-ID extraction, ByteTrack achieves an optimal balance between tracking accuracy and real-time latency.
\section{Variants of YOLO Algorithm for UAV-Based Vehicle Detection}
\label{sec:yolo_variants}
Building upon the general architectural framework established in Section~\ref{sec:architectural_components}, distinct YOLO variants have been developed to address the specific bottlenecks of aerial surveillance. While the generic pipeline provides the foundation, these variants introduce targeted modifications to optimize performance for small object detection and onboard efficiency. This section reviews these notable architectural evolutions, focusing on backbone network replacements, the integration of attention mechanisms, head module decoupling, and the shift towards Transformer-based designs. 

\subsection{Backbone Replacement Variants}
The performance of YOLO is influenced largely by its backbone network that extracts features from input images. Researchers have adopted the better architecture of the original backbone like EfficientNet~\cite{tan2019efficientnet}, MobileNet~\cite{howard2017mobilenets}, and ResNet~\cite{he2016deep} to enhance accuracy and speed. The EfficientNet-based versions employ efficient scaling where the accuracy and efficiency are balanced, thus it is suitable in UAVs using a limited processing power~\cite{mahasin2022comparison}. One of the benefits of MobileNet is that depth wise separable convolution reduces computation depth load, a desired feature when running an application in real time~\cite{jia2023mobilenet}. These backbone enhancements enhance the performance especially with regard to detecting dense and small vehicles in aerial imagery.

\subsection{Attention Mechanisms}
Attention mechanisms have been incorporated into YOLO to enhance its focus on relevant features and suppress background noise. Examples of interest are Squeeze-and-Excitation networks~\cite{hu2018squeeze} and Convolutional Block Attention Modules (CBAM)~\cite{woo2018cbam}. These techniques improve feature maps with the focus on important areas, which matters particularly in UAV imagery when the vehicles merge in a complicated background or are of different scales and orientations. Research indicates that attention-based YOLO variants obtain better precision and recall performances compared with the original, proving the importance of adjustable feature priorities in the complicated tasks of detection.

\subsection{Head Module Enhancements}
The head module of YOLO that makes predictions based on extracted features has been enhanced to be more accurate in detection. Anchor-free designs eliminate the dependence on pre-defined anchor boxes and present ease of use and excellent detection capability on small objects~\cite{tian2022fully}. Also, focal loss and generalized IoU loss functions reduce class imbalance and enhance the accuracy of localization~\cite{lin2017focal, rezatofighi2019generalized}. These improvements increase the reliability of detection even under changing environment conditions.

\subsection{Transformer-Based YOLO and Attention-Enhanced Architectures}
Transformer models have played a major part in the development of YOLO~\cite{carion2020end}. Due to their ability to model long-range dependencies and context much better than CNNs, they make detection more effective in unfavorable settings, like the different orientation of different vehicles and uneven distribution of objects~\cite{zhang2021vit, chen2021you}. Their computational expense has, however, triggered innovations to be in line with the overall trend toward attention - based object detection models~\cite{khan2022transformers}.

YOLOv12 \cite{tian2025yolov12} mitigates this issue of incorporating attention on real-time detection. It preserves YOLO’s balance of latency and accuracy while introducing novel attention-based modules without sacrificing speed. A core innovation, the Area Attention Module (A2), reduces the high computational cost of conventional attention by dividing the feature map into a small number of regions. This design has a broad receptive field and low computational overhead. The Residual Efficient Layer Aggregation Network (R-ELAN) solves the deficiencies of previous feature aggregation networks. The first ELAN was found to have a block of the gradient and stability problems when using large-scale training with attention. R-ELAN overcomes these issues through block-level residual connections with scaling factors, improving gradient flow and convergence. Its redesigned bottleneck-based aggregation integrates features efficiently while reducing computation, parameters, and memory usage. Additional YOLOv12 optimizations include Flash-attention to minimize the inefficiency of memory access, omitting positional encoding to support architectural simplicity, adjusting the ratio of MLP to balance the costs of attention and feedforward, and modifying the stacked-block structure of two R-ELAN blocks to provide only one block towards the bottom of the stack to enhance training performances.

Drone-YOLO \cite{zhang2023drone}, was developed to address UAV-specific challenges such as large images, small objects, dense scenes, and poor lighting. It has a number of innovations that improve small-object recognition at low computational cost. RepVGG modules can be used as down sampling layers in order to enhance multi-scale learning~\cite{ding2021repvgg}. The neck uses a four-layer PAFPN as well as an innovative Sandwich-Fusion (SF) module. The SF module is integrated into every top-down level and combines high-level and low-level features, optimizing spatial and semantic information. Depthwise separable convolution also enhances the feature extraction and increases the receptive field with lower cost.

DAU-YOLO~\cite{zeyu2025dau}, introduces innovations that enhance accuracy while keeping the model lightweight. Its backbone modules include a Receptive-Field Attention (RFA) module, which locally adapts convolution kernels with an aim of separating overlapping objects and feature extractions of small objects. A Dynamic Attention and Up-sampling (DAU) module is used in the neck, which enhances the detection of small objects via the utilization of finer low-level details. For achieving feature refinement, Scale-Diffusion Attention (with deformable convolution) and Task-Aware Attention are used, such that the detection tasks are prioritized dynamically.
\section{Benchmark Datasets and Evaluation Metrics}
\label{sec:benchmark_datasets}
The size, type, and quality of datasets play a critical role in the development of the deep-learning algorithms for vehicle detection, particularly with UAV imagery.  Despite the lack of well-annotated UAVs datasets, comprehensive and challenging datasets will be critical to the field. In this section, some common benchmark datasets and evaluation criteria referred to when evaluating deep-learning-based vehicle detection algorithms are reviewed.

\subsection{Selected Datasets for This Study}
The two datasets used in this study are the VisDrone~\cite{du2019visdrone} and the Aerial Multi-Vehicle Detection Dataset (AMVD)~\cite{rafael_makrigiorgis_2022_7053442}, selected for their diversity, comprehensiveness, and relevance to aerial vehicle detection. Model performance is determined on these datasets to give robust results of aerial traffic monitoring.

\subsubsection{VisDrone Dataset}
The VisDrone Dataset~\cite{du2019visdrone}, developed by the AISKYEYE team at Tianjin University, is a large-scale benchmark for UAV-based computer vision tasks. It includes 288 video clips containing 261,908 frames and 10,209 static images, with more than 2.6 million marked bounding boxes of objects like people, vehicles, bicycles, and tricycles. The data were gathered in 14 cities in China that had different urban and countryside environment, weather conditions, and light conditions. VisDrone dataset supports tasks including object detection, single and multi-object tracking, and crowd counting.

\subsubsection{Aerial Multi-Vehicle Detection Dataset}
The Aerial Multi-Vehicle Detection Dataset~\cite{rafael_makrigiorgis_2022_7053442} is developed for traffic monitoring and vehicle-detecting using UAV imagery. It has 9,048 high-resolution images taken in the busy roadways of Nicosia and Limassol, Cyprus, annotated with 248,413 cars, 2,002 buses and 7,744 trucks. Images were acquired at 150-250 meters to have a uniform top-down view. 

These datasets were chosen because of their complementary strengths and direct applicability to the detection of vehicles in real world traffic scenarios. A key strength of the VisDrone dataset is its diversity, featuring a wide variety of scenes and environmental conditions from several urban cities in China. This heterogeneity is instrumental in training a model that is robust to diverse backgrounds. In contrast, the Aerial Multi-Vehicle Detection Dataset, which focuses on traffic surveillance in Cyprus, is characterized by images captured from a higher altitude, enabling the monitoring of a wider area. Both datasets offer extensive collections of annotated images, which are fundamental for training and evaluating deep learning models. While VisDrone includes a wide range of object categories, the Aerial Multi-Vehicle Detection Dataset is specifically focused on vehicles from a high-altitude perspective. This complementarity allows for a comprehensive assessment of each algorithm's performance and a more rigorous validation of their generalization capabilities across different operational conditions.

In the comparative experiments, different model architectures were evaluated, to identify the optimal balance between detection accuracy and computational efficiency for UAV deployment. Since UAV onboard computers have only limited computational resources, the experiment prioritized the small (s) and medium (m) variants of popular detection architectures. This choice aimed to minimize inference time and model size without compromising detection accuracy. Evaluation metrics included mean Average Precision (mAP) at an Intersection over Union (IoU) threshold of 0.5 (mAP@0.5), class-specific Average Precision (AP) of individual object categories, model parameters (in millions) and Floating-Point Operations (FLOPs) as the indicator of computing complexity.

\subsection{Other Datasets in UAV-based Vehicle Detection}
\subsubsection{pNeuma Dataset}
The pNeuma dataset~\cite{barmpounakis2020new} is a large-scale UAV-based collection containing around 500,000 urban trajectories in 12.5 hours of the highest traffic intensity. It gives thorough lane-detecting and vehicle-maneuvering information that is vital in algorithms that apply to urban traffic conditions. The drones were used to capture the trajectories and all the vehicles were covered within the study area. The trajectories are calibrated against the World Geodetic System (WGS 84) and are captured along with the same frequency of 0.04-seconds computed as the highest frame rate of the video. Besides to position data, it includes derived features such as speed, acceleration, and distance traveled. It also specifies vehicle types, including cars, taxis, motorcycles, buses, and heavy or medium vehicles. 

\subsubsection{NGSIM Dataset}
The NGSIM dataset~\cite{coifman2017critical} provides the trajectory of vehicles on four locations, which include both freeway and arterial segments. It is a universal vehicle-detection and traffic-analysis resource where the major trajectory data as well as a range of additional information is given. Data was obtained through the use of static cameras that were placed along two arterial and two freeway segments. Besides vehicle paths, it includes location-specific data such as ortho-rectified photographs, CAD drawings, signal timings, weather, and detector readings. It can also provide raw and processed video files which make it an even greater asset to research and analytical tasks.

\subsubsection{Stanford Drone Dataset}
The Stanford Drone Dataset (SDD)~\cite{robicquet2016learning} is widely used for analyzing human and vehicle trajectories in crowded urban environments. It includes roughly 9 hours of UAV videos, 10,000 trajectories, and 930,000 samples across eight locations on the Stanford campus. It also contains a large variety of objects: vehicles, pedestrians, cyclists, skateboards, carts, and buses. Remarkably, 7\% of captioned targets are vehicles, and most data are on pedestrians and cyclists, with trajectories mapped in both space and time.

\subsubsection{VSAI Dataset}
The VSAI dataset~\cite{wang2022vsai} is a multi-view UAV dataset which is particularly formulated to attain vehicle detection in complex situations. It consists of 9,000 images having 87,000 annotated objects labeled as either small or large vehicles. The dataset is valuable in training detection algorithms as a platform of testing their efficiency in dealing with the varied and difficult scenarios of aerial imaging situations.

\subsubsection{MONET Dataset}
The MONET dataset~\cite{riz2023monet} is a comprehensive multi-modal dataset captured using a thermal camera mounted on a UAV. It contains 53,000 images with 162,000 annotated bounding boxes, accompanied by rich metadata such as UAV altitude, speed, and GPS coordinates. Collected over rural areas of Nicosia, MONET offers a unique perspective on detecting people and vehicles in aerial thermal imagery. Each image is precisely timestamped and aligned with corresponding drone metadata, offering valuable contextual information for each scene. The dataset contains annotations for both human and vehicle targets, along with trajectory information, enhancing its value for computer vision and robotics applications. By combining thermal imagery, detailed annotations, and extensive metadata, MONET serves as a valuable resource for research in aerial object detection, tracking, and multi-sensor fusion algorithms. Its diverse rural scenes and varied flight conditions enhance its potential for developing robust models for real-world UAV applications.

\subsubsection{UAVDT Benchmark}
The UAVDT (UAV Detection and Tracking) benchmark~\cite{du2018unmanned} is a comprehensive dataset designed for vehicle detection and tracking in urban environments. It comprises 100 video sequences totaling 10 hours of high-quality UAV footage from diverse urban locations, including squares, arterial streets, toll stations, highways, intersections, and T-junctions. From this collection, 80,000 selected frames are provided, each annotated with 14 key attributes. These attributes encompass environmental and operational factors such as weather, UAV altitude, and camera angle, along with target-specific details like vehicle type and occlusion level. By encompassing diverse real-world scenarios, this dataset supports the development and evaluation of robust models capable of performing effectively in varied urban environments.

\subsubsection{AU-AIR Dataset}
The AU-AIR dataset~\cite{bozcan2020air} is a comprehensive multi-modal sensor data set capture during low- altitude UAV flights (10-30m). It includes 2 hours of videos, generating 32,000 images that were captured in a variety of lighting and weather conditions. The multi-modal observations have visual and temporal data, geo-spatial (longitude/latitude), and altitude information, velocity and UAV orientation parameters (roll, pitch, and yaw). Data collection using a low-altitude level increases the relevance of the data set in the urban and near-ground aerial scene surveillance.

\subsubsection{AD4CHE Dataset}
The AD4CHE dataset~\cite{zhang2023ad4che} provides an extensive collection of aerial survey data that includes 5 hours of recordings of four Chinese mega-cities. Notably, this dataset focus on peak traffic hours and includes trajectory information alongside the digital maps. AD4CHE allows particular insight into urban mobility and traffic patterns through its urban diversity and time-centric approach, which makes it valuable in research within highly populated cities.

\subsubsection{AUTOMATUM Dataset}
The AUTOMATUM dataset~\cite{spannaus2021automatum} provides 30 hours of drone data consisting of 12 highway-like scenes. The main attribute is its focus on the generation of object trajectories or connecting the recognized objects with their coordinate. The method makes it possible to carry out exact spatial analysis and study behavior of vehicles on highways.

\subsubsection{SIND Drone Dataset}
The SIND Drone dataset~\cite{xu2022drone} provides 7 hours of recordings tracking 13,000 participants in traffic at urban intersections. It is characterized by detailed annotations making information available on trajectories, motion states (position, velocity, acceleration, heading direction, yaw angle), traffic light states for eight signals. It also includes motion parameters, high resolution maps, and metadata regarding different tracks. Traffic participants are sorted into seven categories cars, bus, truck, motorcycle, bicycle, tricycle, and pedestrian, which gives a clear picture of the dynamics of urban traffic.

\subsubsection{INTERACTION Dataset}
The INTERACTION dataset~\cite{zhan2019interaction} provides more naturalistic motion measurements of those present in the road traffic under highly interactive driving conditions. It covers several nations such as China, Germany, and the United States, and this offers quite varied geographic and cultural backgrounds. It includes all road topographies: roundabouts, signalized intersections, signalized intersections, as well as ramp merging or lane changing maneuvers. One relevant aspect of this data source is the availability of maps with so-called raster semantics, such as physical layers, reference lines, lanelet connections in addition to the traffic rules. The recorded trajectory information can provide great value in the study of the complex traffic style in support of autonomous vehicle research and traffic flow analysis.

\subsubsection{STVD Dataset}
The Spatio-Temporal Vehicle Detection (STVD) dataset~\cite{telegraph2024spatiotemporal} is designed for spatiotemporal object detection in UAV-based aerial imagery. Unlike datasets based on single-area, low-resolution satellite imagery, STVD is built from multiple aerial video clips of traffic captured in different road segments of Nicosia, Cyprus. Extracting multiple image sequences from these videos yields about 6,600 high-resolution (1920$\times$1080) frames. A notable feature of STVD is its spatiotemporal consistency, with frames from the same continuous sequence preserved. This supports the development of detection methods that exploit multiple sequential frames to improve accuracy, offering a balanced setup for both standard object detection and advanced temporal modeling.

\begin{table*}[ht!]
\centering
\renewcommand{\arraystretch}{1.2}
\caption{Comparison of various datasets used in UAV-based vehicle monitoring (VD: Vehicle Detection, VC: Vehicle Classification, VT: Vehicle Tracking, SE: Speed Estimation, VCN: Vehicle Counting)}
\label{tab:datasets}
\begin{tabularx}{\textwidth}{@{} l l l l l X l l @{}}
\toprule
\textbf{Dataset} & \textbf{View} & \textbf{Task} & \textbf{Platform} & \textbf{Altitude (m)} & \textbf{Data Volume} & \textbf{Number of Instances} & \textbf{Image Size} \\
\midrule

Stanford Drone~\cite{robicquet2016learning} & Top & VD, VCN & UAV & $\sim$80 & 60 videos & 930K & $1400\times1904$ \\

UAVDT~\cite{du2018unmanned} & Multi & VD, VT & UAV & 10--70+ & 100 videos, 80K frames & 840K+ & $1080\times540$ \\

AU-AIR~\cite{bozcan2020air} & Various & VD & UAV & 10--30 & 2h video, 32K frames & - & $1920\times1080$ \\

DOTA~\cite{xia2018dota} & Various & VD & Aerial & - & 2,806 images & 188,282 & $4000\times4080$ \\

UAV-ROD~\cite{zhou2022ts4net} & Various & VD, SE, VCN & UAV & 30--80 & 1,576 images & - & \makecell[l]{$1920\times1080$ to\\$2720\times1530$} \\

DIOR-R~\cite{cheng2022anchor} & Top & VD & Aerial & - & 23,463 images & 192,512 & $800\times800$ \\

SODA-A~\cite{cheng2023towards} & Top & VD & Aerial & - & 2,513 images & 872,069 & $4761\times2777$ \\

MOR-UAV~\cite{mandal2020mor} & Various & VD & Aerial & - & 10,948 images & 89,783 & \makecell[l]{$1280\times720$ to\\$1920\times1080$} \\

VisDrone~\cite{du2019visdrone} & Various & VD & UAV & - & 10k images, 263 videos & 2.5 M & $2000\times1500$ \\

STVD~\cite{telegraph2024spatiotemporal} & Various & VD, VT & UAV & - & 6,600 images & 84,331 & $1920\times1080$ \\ 

AMVD~\cite{rafael_makrigiorgis_2022_7053442} & Various & VD, VT & UAV & 150--250 & 9,048 images & 258,159 & \makecell[l]{$1920\times1080$ to\\$3840\times2160$} \\

\bottomrule
\end{tabularx}
\end{table*}

\subsection{Limitations of Existing Datasets and Future Needs} 
While numerous datasets have advanced the field of UAV-based vehicle detection, several critical limitations remain that hinder full-scale real-world deployment.

\subsubsection{Limited Environmental Diversity} 
Most existing datasets are predominantly captured under favorable weather conditions. There is a significant scarcity of annotated data covering adverse weather scenarios such as heavy rain, fog, snow, or night-time conditions. This data gap limits the ability of current models to generalize to all-weather surveillance tasks, which is a prerequisite for 24/7 traffic monitoring systems.

\subsubsection{Lack of Holistic Annotations} 
Current benchmarks typically focus on a single task, isolating either detection (bounding boxes) or tracking (IDs). Comprehensive datasets that simultaneously provide multi-modal annotations to combine detection, pixel-level segmentation, trajectory behavior, and 3D orientation are rare. Such holistic data is essential for developing "end-to-end" perception systems that can not only detect a car but also understand its precise lane position and intent.

\subsubsection{Geographical and Cultural Bias} 
A large portion of high-quality aerial datasets originates from highly specific regions, which introduces geographical and cultural bias. For instance, widely used benchmarks such as VisDrone are predominantly collected in Chinese urban centers the Stanford Drone Dataset is restricted to a single university campus in the United States. Since traffic patterns, vehicle types, and road infrastructures vary significantly across continents, this lack of a globally diverse dataset covering multiple countries and traffic rules restricts the cross-domain generalization of trained models.

To bridge these gaps, future data collection initiatives must prioritize meteorological diversity by systematically capturing adverse weather scenarios. Furthermore, enriching benchmarks with synchronized multi-modal data is essential to advance robust sensor fusion research. Ultimately, fostering global collaboration to aggregate geographically diverse annotations will be pivotal in establishing a truly universal standard for UAV-based traffic monitoring.

\begin{table*}[ht!]
\centering
\caption{Results for individual categories on the VisDrone2019-val Dataset.}
\begin{tabular}{lccccccccccc}
\hline
\textbf{Method} & \textbf{Pedestrian} & \textbf{Person} & \textbf{Bicycle} & \textbf{Car} & \textbf{Van} & \textbf{Truck} & \textbf{Tricycle} & \textbf{Awning-Tricycle} & \textbf{Bus} & \textbf{Motor} & \textbf{All}\\
\hline
Faster R-CNN \cite{ren2016faster} & 21.4 & 15.6 & 6.7 & 51.7 & 29.5 & 19.0 & 13.1 & 7.7 & 31.4 & 20.7 & 21.7 \\
Cascade R-CNN \cite{cai2018cascade} & 22.2 & 14.8 & 7.6 & 54.6 & 31.5 & 21.6 & 14.8 & 8.6 & 34.9 & 21.4 & 23.2\\
DMNet \cite{li2020dmnet} & 28.5 & 20.4 & 15.9 & 56.8 & 37.9 & 30.1 & 22.6 & 14.0 & 47.1 & 29.2 & 30.3\\
YOLOv3(lite) & 34.5 & 23.4 & 7.9 & 70.8 & 31.3 & 21.9 & 15.3 & 6.2 & 40.9 & 32.7 & 28.5\\
CDNet \cite{wang2014cdnet} & 35.6 & 19.2 & 13.8 & 55.8 & 42.1 & 38.2 & 33.0 & 25.4 & 49.5 & 29.3 & 34.2\\
YOLOv5(s) & 35.8 & 30.5 & 10.1 & 65.0 & 31.5 & 29.5 & 20.6 & 11.1 & 41.1 & 35.4 & 31.1\\
YOLOv5(m) & 41.7 & 34.6 & 14.3 & 71.7 & 38.7 & 36.3 & 24.0 & 12.7 & 47.6 & 40.6 & 36.2\\
MSA-YOLO \cite{su2023msa} & 33.4 & 17.3 & 11.2 & 76.8 & 41.5 & 41.4 & 14.8 & 18.4 & 60.9 & 31.0 & 34.7\\
YOLOv7(tiny) & 38.9 & 35.5 & 9.6 & 76.2 & 38.0 & 28.8 & 20.6 & 10.9 & 48.7 & 43.2 & 35.0\\
YOLOv7(x) & 45.1 & 37.0 & 16.8 & 75.0 & 41.2 & 39.0 & 27.3 & 13.5 & 50.3 & 42.8 & 38.8\\
YOLOv8(m) & 47.1 & 37.4 & 19.9 & 80.1 & 46.9 & 43.4 & 32.8 & 18.2 & 62.2 & 48.8 & 43.7\\
YOLOv9(c)  & 46.5 & 35.8 & 18.6 & 81.2 & 50.7 & 45.6 & 35.8 & 19.0 & 62.9 & 49.7 & 44.6\\
YOLOv11(m)  & 51.1 & 38.3 & 20.4 & 83.0 & 49.9 & 45.9 & 35.5 & 21.7 & 65.3 & 51.6 & 46.3\\
YOLOv12(m)  & 51.0 & 38.1 & 20.2 & 82.8 & 49.2 & 45.1 & 34.4 & 19.5 & 65.3 & 51.1 & 45.7\\
YOLOv26(m) & 56.5 & 42.8 & 21.3 & 84.3 & 48.5 & 44.6 & 35.6 & 23.1 & 66.6 & 55.1 & 47.8 \\
\hline
\label{tab:VisDrone-individual-categories}
\end{tabular}
\end{table*}

\begin{table*}[ht!]
\centering
\caption{Comparison of different models using the VisDrone2019-val dataset.}
\begin{tabular}{p{1.5cm} 
                >{\centering\arraybackslash}p{1.5cm} 
                >{\centering\arraybackslash}p{2.0cm} 
                >{\centering\arraybackslash}p{1.5cm} 
                >{\centering\arraybackslash}p{1.5cm} 
                >{\centering\arraybackslash}p{2.0cm} 
                >{\centering\arraybackslash}p{1.0cm} 
                >{\centering\arraybackslash}p{2.5cm}}
\hline
\textbf{Method} & \textbf{mAP@0.5} & \textbf{mAP@0.5:0.95} & \textbf{Params (M)} & \textbf{FLOPS (G)} & \textbf{Inference (ms)} & \textbf{FPS} & \textbf{Model Size (MB)} \\
\hline
YOLOv5(m)       & 36.2 & 20.8 & 20.9  & 48.0 & 6.2 & 161 & 40.8 \\
YOLOv8(s)       & 43.0 & 26.0 & 11.1  & 28.5 & 4.1 & 245 & 21.5 \\
YOLOv8(m)       & 44.5 & 27.1 & 25.8  & 78.7 & 5.0 & 200 & 49.7 \\
YOLOv9(s)       & 42.1 & 25.3 & 7.2  & 26.7 & 8.6 & 116 & 14.3 \\
YOLOv9(m)       & 45.0 & 27.4 & 20.0  & 76.5 & 8.1 & 123 & 38.7 \\
YOLOv10(s)       & 41.2 & 24.8 & 8.0  & 24.5 & 5.1 & 196 & 31.4 \\
YOLOv10(m)       & 45.3 & 28.1 & 16.5  & 63.5 & 6.1 & 164 & 63.8 \\
YOLOv11(s)      & 41.6 & 25.2 & 9.4  & 21.3 & 4.4 & 228 & 18.4 \\
YOLOv11(m)      & 46.3 & 28.8 & 20.0  & 67.7 & 5.2 & 192 & 38.7 \\
YOLOv12(m)      & 45.7 & 28.4 & 20.1  & 67.2 & 5.6 & 161 & 38.0 \\
YOLOv26(m)       & 47.8 & 29.4 & 20.4  & 67.9 & 4.6 & 217 & 44.1 \\
\hline
\label{tab:VisDrone-results}
\end{tabular}
\end{table*}

\begin{table*}[ht!]
\centering
\caption{Comparison of experimental results on the Aerial Multi-Vehicle Detection dataset.}
\begin{tabular}{lcccccccccc} 
\hline
\textbf{Method} & \textbf{Car} & \textbf{Bus} & \textbf{Truck} & \textbf{mAP@0.5} & \textbf{Params} & \textbf{FLOPS} & \textbf{Inference (ms)} & \textbf{FPS} & \textbf{Precision} & \textbf{Model Size (MB)}\\
\hline
RT-DETR(l)      & 95.2 & 89.7 & 83.4 & 89.4 & 31.9M & 103.4G & 85.7 & 12 & 0.89 & 122.3 \\
YOLOv5(m)       & 97.6 & 93.2 & 93.1 & 94.6 & 20.8M & 47.9G  & 4.3  & 233 & 0.93 & 75.5 \\
YOLOv7          & 96.9 & 93.0 & 92.6 & 94.2 & 36.5M & 103.2G & 8.2  & 122 & 0.92 & 131.4 \\
YOLOv8(m)       & 96.6 & 92.3 & 92.3 & 93.7 & 25.8M & 78.7G  & 12.3 & 81  & 0.91 & 94.0 \\
YOLOv9(c)       & 97.0 & 92.8 & 90.2 & 90.2 & 25.4M & 102.5G & 6.6  & 151 & 0.91 & 90.2 \\
YOLOv10(n)      & 94.6 & 89.8 & 87.5 & 90.6 & 2.7M  & 8.2G   & 7.0  & 143 & 0.87 & 10.2 \\
YOLOv10(m)      & 95.8 & 93.6 & 91.7 & 93.7 & 16.4M & 63.4G  & 13.0 & 77  & 0.91 & 61.8 \\
YOLOv11(n)      & 96.9 & 94.1 & 91.0 & 94.0 & 2.6M  & 6.3G   & 0.9  & 1111 & 0.93 & 9.5 \\
YOLOv11(m)      & 97.7 & 94.3 & 94.3 & 95.1 & 20.0M & 67.7G  & 5.2  & 192 & 0.94 & 72.4 \\
YOLOv12(m)      & 97.6 & 93.4 & 92.8 & 94.6 & 20.1M & 67.1G  & 6.2  & 161 & 0.90 & 40.8 \\
YOLOv26(m)      & 98.5 & 94.4 & 94.2 & 95.7 & 20.4M & 67.9G  & 5.3  & 189 & 0.93 & 42.0 \\
\hline
\label{tab:AMV-results}
\end{tabular}
\end{table*}
\section{Comparative Experiments}
\label{sec:comparative_experiments}

\subsection{Experimental setup}
To ensure experimental consistency, all models were trained with PyTorch 1.12.1, using a standardized hardware platform comprising two Intel Xeon Gold 6240 CPUs, four Nvidia Tesla V100-SXM2-32GB GPUs, and 377 GB RAM. The system was running on AlmaLinux 9.4. The software environment included CUDA 12.6, cuDNN 9.3.0, and OpenCV 4.11.0. To maintain a uniform baseline for comparison and balance computational efficiency with detection performance, all input images were resized to a standardized resolution of $640 \times 640$ pixels during both training and inference. Image augmentation was also applied to improve model generalization.

For the edge deployment and energy efficiency benchmarking, experiments were performed on the NVIDIA Jetson Orin NX (16GB) embedded platform. This device operated in a Linux environment (GCC 11.2.0) running Python 3.10.19. The inference software stack utilized PyTorch 2.5.0 (build nv24.08) and Torchvision 0.20.0. This dual-platform setup ensured that performance differences arose solely from model architecture variations rather than hardware disparities during training, while providing realistic latency and energy metrics for deployment.

\begin{table*}[ht]
\centering
\caption{Inference Performance and Energy Efficiency Analysis on Nvidia Jetson Orin NX (16GB) under Different Power Modes.}
\label{tab:power_efficiency}
\resizebox{\textwidth}{!}{%
\begin{tabular}{llcccccc}
\toprule
\textbf{Model} & \textbf{Power Mode} & \textbf{Power Limit (W)} & \textbf{mAP@0.5} & \textbf{mAP@0.5:0.95} & \textbf{Inference (ms)} & \textbf{FPS} & \textbf{Energy (J/Frame)} \\
\midrule
\multirow{4}{*}{YOLO11n} 
 & Low Power       & 10W & \multirow{4}{*}{37.6} & \multirow{4}{*}{23.8} & 22.9 & 43.7 & \textbf{0.229} \\
 & Balanced        & 15W &                       &                       & 23.1 & 43.3 & 0.346 \\
 & Standard        & 25W &                       &                       & 20.2 & 49.5 & 0.500 \\
 & High Performance & 40W &                       &                       & \textbf{10.7} & \textbf{93.5} & 0.428 \\
\midrule
\multirow{4}{*}{YOLO11s} 
 & Low Power       & 10W & \multirow{4}{*}{41.6} & \multirow{4}{*}{25.2} & 40.8 & 24.5 & \textbf{0.408} \\
 & Balanced        & 15W &                       &                       & 39.0 & 25.6 & 0.586 \\
 & Standard        & 25W &                       &                       & 35.7 & 28.0 & 0.893 \\
 & High Performance & 40W &                       &                       & \textbf{15.9} & \textbf{62.9} & 0.636 \\
\midrule
\multirow{4}{*}{YOLO11m} 
 & Low Power       & 10W & \multirow{4}{*}{46.3} & \multirow{4}{*}{28.8} & 86.6 & 11.5 & \textbf{0.870} \\
 & Balanced        & 15W &                       &                       & 83.5 & 12.0 & 1.250 \\
 & Standard        & 25W &                       &                       & 81.1 & 12.3 & 2.032 \\
 & High Performance & 40W &                       &                       & \textbf{33.9} & \textbf{29.5} & 1.356 \\
\bottomrule
\multicolumn{7}{l}{\footnotesize All experiments were conducted using the \textbf{VisDrone2019-val} dataset~\cite{du2019visdrone}.}
\end{tabular}%
}
\end{table*}

\subsection{Experimental Results of the VisDrone Dataset}
The VisDrone dataset was used as the primary benchmark for detection performance. Table~\ref{tab:VisDrone-individual-categories} summarizes the results. YOLOv11m achieved the highest mAP@0.5 at 46.3\%, marking a substantial improvement over earlier versions. Across YOLO versions 3 to 12, performance improved notably, especially for challenging categories such as “Bicycle” and “Awning-Tricycle”. Medium-sized models like YOLOv11m and YOLOv12m offered an optimal trade-off between accuracy and computational cost, making them suitable for UAV deployment where both factors are critical. Our experiments results align with previous work on~\cite{wang2025cf, chen2024sl, muzammul2024enhancing}.

\subsection{Experimental Results of the Aerial Multi-Vehicle Detection Dataset}
To validate the generalization capabilities of the proposed models, further evaluations were conducted on the Aerial Multi-Vehicle Detection dataset, which specifically emphasizes high-altitude, top-down vehicle monitoring characterized by a wide-area footprint. Table~\ref{tab:AMV-results} details the comparative analysis regarding accuracy, parameter count, and computational complexity. While the established YOLOv5m provided a strong baseline with 94.6\% mAP@0.5, the YOLOv11m architecture emerged as the superior candidate, attaining the highest accuracy of 95.1\% mAP@0.5 while maintaining efficiency (20.0 M parameters, 67.7 GFLOPs). Conversely, for scenarios imposing strict hardware limitations, the YOLOv10 nano variant (YOLOv10n) demonstrated exceptional resource efficiency, achieving a competitive 90.6\% mAP@0.5 with minimal computational demand. 

\subsection{Energy Efficiency and Real-time Deployment Analysis}
Beyond standard accuracy metrics, the practical viability of deep learning models on UAVs is fundamentally governed by their energy efficiency and inference latency. To quantify these operational parameters, we conducted extensive benchmarking on the NVIDIA Jetson Orin NX (16GB) embedded platform across four distinct power modes: Low Power (10W), Balanced (15W), Standard (25W), and High Performance (40W). Adhering to the edge benchmarking protocols outlined by Bianco et al.~\cite{bianco2018benchmark}, we measured the inference latency and throughput (FPS) for each model variant. Subsequently, to evaluate the energy cost per inference, we derived the energy-per-frame metric ($E_f$), calculated as the ratio of the power limit to the inference speed ($E_f = P_{limit} / FPS$). This metric serves as a direct indicator of the battery consumption required to process a single image.

The quantitative results, presented in Table~\ref{tab:power_efficiency}, highlight distinct performance characteristics across the power spectrum. The lightweight YOLO11n exhibited the highest energy efficiency, achieving a minimal consumption of 0.229 J/Frame in the 10W mode while maintaining a real-time speed of 43.7 FPS. In the High Performance 40W mode, this model scaled to 93.5 FPS, demonstrating significant throughput potential. Conversely, the larger YOLO11m architecture, while offering higher feature extraction capabilities, incurred a substantially higher computational cost, recording an energy expenditure of 1.356 J/Frame in the 40W mode. 

\subsection{Discussion and Insights}
Our evaluation shows that medium-sized models, particularly YOLOv11m and YOLOv12m, strike an optimal balance between detection accuracy and computational efficiency for aerial vehicle detection. They consistently achieve high mAP scores across both datasets while maintaining manageable parameters and FLOPs. The evolution from YOLOv5 to YOLOv12 brought significant gains, especially for challenging categories; in the VisDrone dataset, newer models excelled at detecting small and occluded objects. 

For practical UAV deployment, our experiments on the NVIDIA Jetson Orin NX further highlight the critical trade-off between performance and energy consumption. While medium models (e.g., YOLOv11m) deliver high accuracy, they demand substantial power budgets, consuming approximately 1.36 J/Frame in high-performance (40W) modes to maintain real-time throughput. In contrast, in highly resource-limited scenarios, lighter models such as YOLOv11n prove to be exceptionally efficient options. Our results demonstrate that the nano variant can sustain real-time detection at just 0.229 J/Frame under a strict 10W power limit, offering reduced computational demand at the cost of slight accuracy. This bifurcation suggests that future systems should select model scales based on specific mission endurance requirements, prioritizing nano models for long-range patrols and medium models for precision-critical tasks.
\section{Open Challenges in Real-world Deployment}
\label{sec:challenges}
Building upon the foundational frameworks of UAV-enabled ITS and the specialized domain of deep learning-based vehicle detection, this section synthesizes the primary challenges and provides an overview of current open problems in UAV-based traffic monitoring.

\begin{table*}[ht]
\centering
\caption{Summary of Key Technologies, Challenges, and Insights in UAV-aided ITS Components.}
\label{tab:lessons_learned}
\resizebox{\textwidth}{!}{%
\begin{tabular}{p{0.12\textwidth} p{0.25\textwidth} p{0.25\textwidth} p{0.32\textwidth}}
\toprule
\textbf{Component} & \textbf{Dominant Approaches} & \textbf{Key Challenges} & \textbf{Insights} \\
\midrule
\textbf{Monitoring} & 
\begin{itemize}
    \item Fixed-path patrol
    \item Hover-and-stare
    \item Multi-UAV coordination
\end{itemize} & 
\begin{itemize}
    \item Limited battery endurance
    \item Restricted field of view (FoV)
    \item Vibration and stability
\end{itemize} & 
Single-UAV systems are insufficient for large-scale monitoring. \textbf{Cooperative swarms} and \textbf{tethered UAVs} are essential for continuous, wide-area coverage. Integration with stationary infrastructure (RSUs) enhances reliability. \\
\midrule
\textbf{Detection} & 
\begin{itemize}
    \item One-stage detectors (YOLO series)
    \item Transformer-based (RT-DETR)
    \item Two-stage detectors (Faster R-CNN)
\end{itemize} & 
\begin{itemize}
    \item Small object scale
    \item High-density occlusion
    \item Varying viewpoints
\end{itemize} & 
Accuracy alone is insufficient; \textbf{SWaP (Size, Weight, and Power)} constraints are critical. Nano-scale models (e.g., YOLOv11n) offer the best trade-off for onboard edge computing, while larger models require ground station offloading. \\
\midrule
\textbf{Tracking} & 
\begin{itemize}
    \item Detection-based (DeepSORT, ByteTrack)
    \item Motion-based (Kalman Filter)
    \item Correlation Filters (KCF)
\end{itemize} & 
\begin{itemize}
    \item ID switching due to occlusion
    \item Abrupt camera motion
    \item Real-time latency
\end{itemize} & 
Tracking robustness relies heavily on detection quality. Algorithms must compensate for \textbf{camera motion (Ego-motion)} explicitly. Lightweight association methods (e.g., ByteTrack) are preferred over complex appearance-based re-ID for real-time UAV applications. \\
\midrule
\textbf{Data Fusion} & 
\begin{itemize}
    \item Early fusion (Pixel-level)
    \item Late fusion (Decision-level)
    \item Multi-view geometry
\end{itemize} & 
\begin{itemize}
    \item Synchronization latency
    \item Bandwidth bottlenecks
    \item Registration errors
\end{itemize} & 
Late fusion is more practical for Multi-UAV systems due to communication constraints. \textbf{Edge-cloud collaboration} is necessary to handle the heavy computational load of fusing high-resolution streams from multiple aerial sources. \\
\bottomrule
\end{tabular}%
}
\end{table*}

\subsection{Challenges in Vehicle Detection from UAV Imagery}
The detection of vehicles in UAV imagery is challenging because of the variations in altitude, viewing angles, and environmental conditions. The subsections below outline the primary challenges in UAV-based vehicle detection.

\subsubsection{Scale Diversity}
In UAV images, object size differs significantly with capture altitude and it is therefore hard to track due to objects being imaged at different scale. The image resolution and data consistency also changes with changes in altitude and this necessitates adaptive frameworks to keep accuracy at different altitudes~\cite{li2023developing}. This problem was not always handled with traditional approaches, which have a set window size in feature extraction. The recent methods, particularly Feature Pyramid Networks (FPN) enhance multi-scale detection, yet tracking small vehicles in high-resolution images are still an open challenge~\cite{lin2017feature}.

\subsubsection{Small Object Detection}
In high altitudes the vehicles take up a minimal part of the image, and it is hard to extract features because essential data is missing. Traditional hand-crafted methods frequently missed such small objects. In spite of the fact that deep learning detectors outperform, they still struggle with limited detail, which often has a higher false-negative rate~\cite{ozbilge2024ensembling, huang2022ufpmp}.

\subsubsection{Vehicle Orientations and Types}
The UAV platforms obtain images in perspectives that include top-down, side and oblique imaging that give the vehicle non-uniform orientations. Angular variations make different shaped and varying types of vehicles differently noticeable, making them hard to detect. Earlier approaches were not able to perform well with change of orientation whereas newer architectures learn more generalized features under diverse viewpoints. However, the problem is in the variability of orientation~\cite{wang2019orientation, xu2017car}.

\begin{figure}[htbp]
    \centering
    \includegraphics[width=\linewidth]{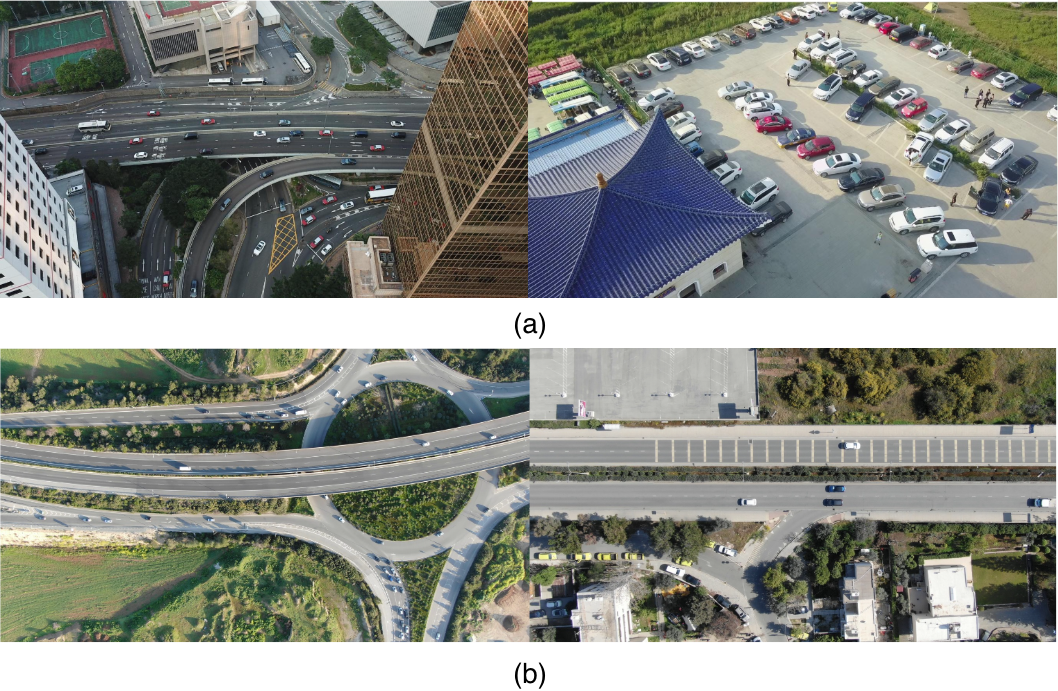}
    \caption{(a) Examples images from the VisDrone Dataset \cite{du2019visdrone}, this collection includes 288 video clips, totaling 261,908 frames, along with 10,209 static images, captured by various drone-mounted camera systems. The dataset covers 14 diverse Chinese cities, including both urban and rural environments, with a wide array of objects such as pedestrians, vehicles, and bicycles in scenes of varying density. (b) Example images from the Aerial Multi-Vehicle Detection Dataset~\cite{rafael_makrigiorgis_2022_7053442}. The dataset comprises 9,048 images captured over busy road segments in Cyprus, with meticulous annotations for three vehicle classes: cars, buses, and trucks. Images, originally in Full HD to 4K resolution, were captured at altitudes of 150–250 meters, providing a consistent top-down perspective.}
    \label{fig:aerial_datasets}
\end{figure}

\begin{figure}[htbp]
    \centering
    \includegraphics[width=0.9\linewidth]{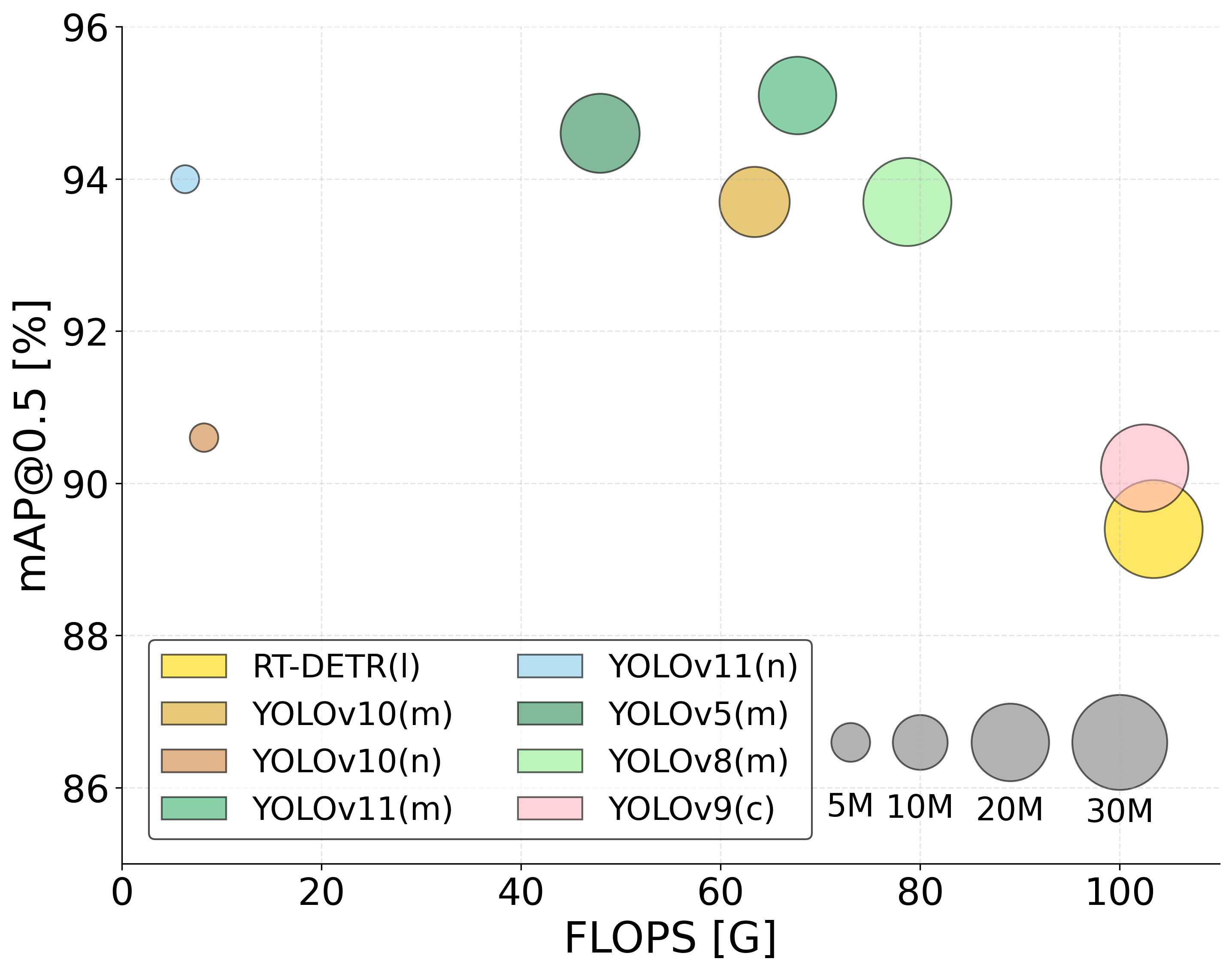}
    \caption{Comparison of experimental results on the Aerial Multi-Vehicle Detection dataset.}
    \label{fig:Comparision_Multi-Vehicle}
\end{figure}

\begin{figure}[htbp]
    \centering
    \includegraphics[width=\linewidth]{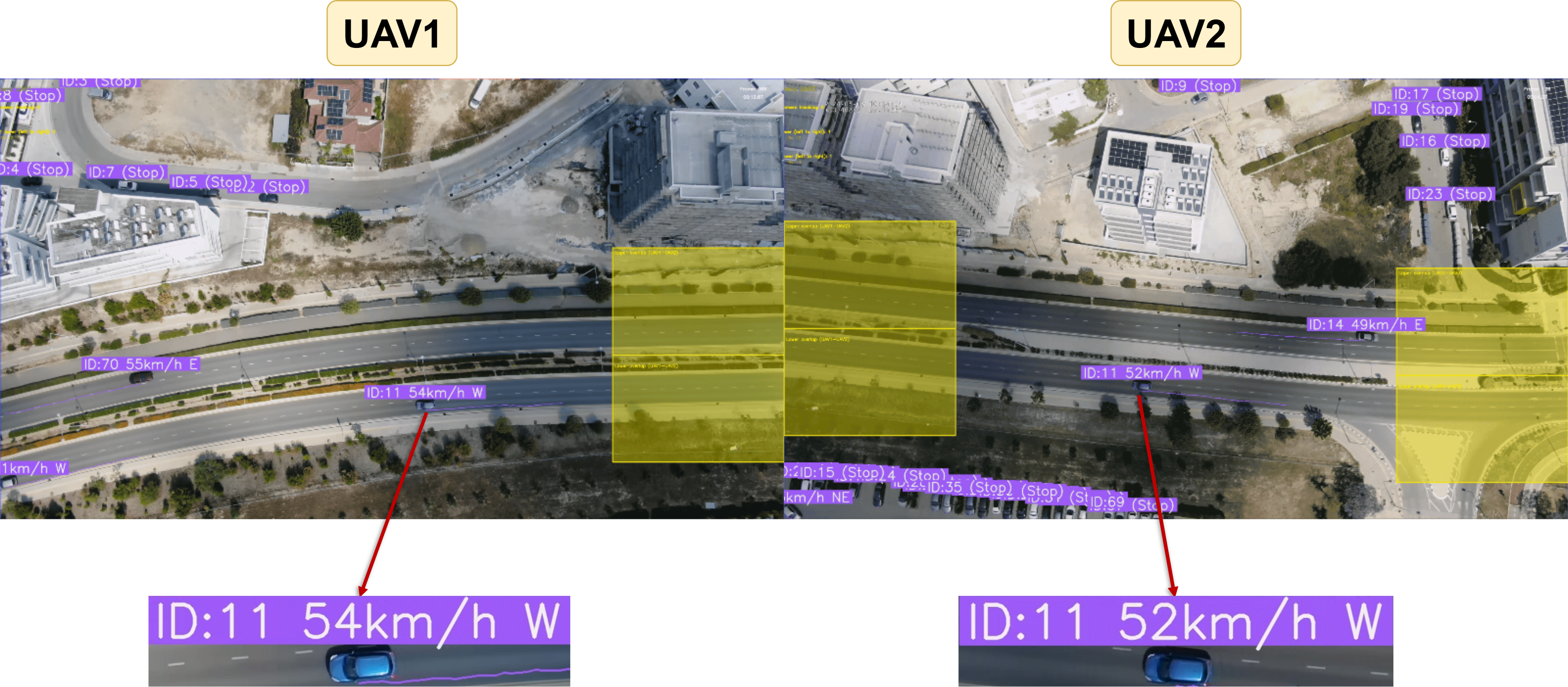}
    \caption{Illustration of multi-camera multi-vehicle tracking in UAV-based traffic monitoring. The yellow regions denote overlapping areas between adjacent UAVs. A vehicle is continuously tracked across different UAV perspectives using re-identification (Re-ID), ensuring that the tracking ID remains consistent. This enables post-processing to directly extract the complete trajectory of the same vehicle for further analysis.}
    \label{fig:Cross_cameras_tracking}
\end{figure}

\begin{figure}[htbp]
    \centering
    \includegraphics[width=\linewidth]{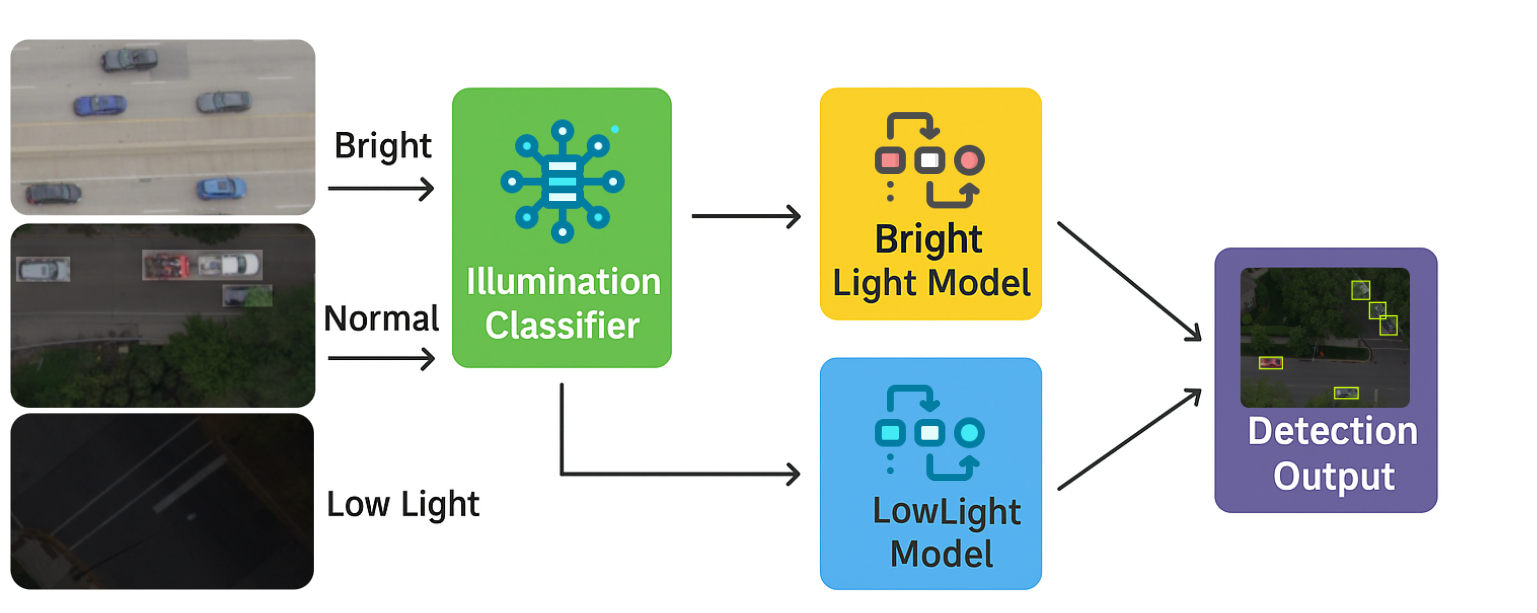}
    \caption{Model selection of vehicle detection based on different illumination environments.}
    \label{fig:Illumination}
\end{figure}

\subsubsection{Occlusion and Environmental Dynamics}
In UAV based monitoring, occlusion is a major challenge, particularly where vehicles may be obstructed by other infrastructure, pedestrians or other vehicles in dense urban traffic. These occlusions disrupt the continuity of the data and need deeper post-processing to deduce any missing spatial-temporal data~\cite{telegraph2024spatiotemporal}. In addition to occlusion, environmental dynamics make detection less reliable as well:

\begin{itemize}
\item \textbf{Illumination Variations}: The appearance of an object can be greatly varied by the changes in lighting caused by time of day, weather or shadows. As an example, the sunlight or low-light at dusk may lead to obscuring vehicle features, and shadows form false edges that confuse detectors (see Figure~\ref{fig:Illumination}). While traditional approaches were exceptionally sensitive to such variations, but in the contemporary practices, tools such as data augmentation and image normalization enhance their robustness.
\item \textbf{Adverse Weather}: Rain, fog, and snow decrease the level of image clarity as well as image contrast, degrading the effectiveness of detectors using either traditional or deep learning approaches. 
\end{itemize}

These factors require adaptive approaches that are capable of dealing effectively with changing environmental conditions, while also offering real-time processing speed. Furthermore, recent studies highlight the necessity of moving beyond single-platform solutions to ensure continuous operation under adverse weather. As demonstrated in state-of-the-art architectures, deploying UAV-UGV cooperative systems offers a highly robust alternative. In such configurations, ground vehicles (UGVs) not only provide supplementary viewpoints when aerial visual reliability is degraded by inclement weather, but also serve as mobile energy management nodes, thereby sustaining long-term, dynamic urban monitoring missions~\cite{oubbati2025uav}.

\subsubsection{Background Complexity and Object Similarity}
Wide-angle UAV imagery often has complex, cluttered backgrounds. Vehicles can be blocked out or mistaken as similar-looking objects such as rubbish bins or air conditioning units. Traditional methods had difficulty identifying cars in these situations. Current CNN detectors have lowered false positives by employing better feature extraction, but object similarity and background complexity continues to be a significant challenge~\cite{girshick2015fast, ren2016faster}.

\subsubsection{Limited comprehensive Datasets}
A key limitation in UAV vehicle detection is the lack of large dataset with high-quality annotations. Most available datasets are from satellites or ground-based cameras, which is vastly different from UAV-captured images, resulting in a lack of optimal performance when using pre-trained models. More comprehensive datasets that include a wide variety of traffic scenarios are required to improve the performance. Specifically, these datasets should capture diverse road environments, a range of weather conditions, and a sufficient variety of vehicle types and classes.

\subsubsection{Real-Time Processing Constraints}
Real-time vehicle detection is essential for UAV applications such as traffic monitoring and search-and-rescue. However, complex deep learning models demand significant computational resources, often exceeding UAV hardware capabilities. Lightweight architectures like YOLOv4 and MobileNet address this issue, yet achieving a balance between accuracy and speed remains an active research challenge.

Real-time vehicle detection is crucial for UAV applications such as traffic monitoring and search-and-rescue. The deeper models of deep learning, however, are both computationally resource intensive and usually surpass the hardware provisions of UAVs. This can be address by lightweight architectures such as the ShuffleNet and MobileNet~\cite{zhang2018shufflenet, sandler2018mobilenetv2}. However, a balance between accuracy and speed is an open research problem. 

\subsection{Traditional Methods in Vehicle Detection}
Background subtraction, edge detection and optical flow are some of the traditional techniques of vehicle detection. Such methods exploit manually coded features and rule-based classifies. To illustrate, vehicle detection using shape and texture has been done using Histogram of Oriented Gradients with Support Vector Machines~\cite{dalal2005histograms}. However, these techniques tend to be non-generalizing in complex settings and are computationally intensive, making real-time deployment challenging.

\subsection{Deep Learning Methods in Vehicle Detection}
Advances in deep learning, along with improved processing power, have made the processing speed of object detection much faster. CNNs have transformed the field through their ability to extract complicated and complex features automatically, outperforming traditional methods~\cite{girshick2014rich}. By eliminating the need for manual feature engineering, CNNs significantly reduce human bias and improve overall accuracy. Given this powerful hierarchical feature extraction capability, most contemporary UAV vehicle detection frameworks are fundamentally based on CNN models~\cite{he2015spatial}.

These models are widely divided into two types: two-stage detectors and one-stage detectors. Two-stage detectors, or region-based approaches, include R-CNN~\cite{girshick2014rich}, SPPnet~\cite{he2015spatial}, Fast R-CNN~\cite{girshick2015fast}, Faster R-CNN~\cite{ren2016faster}, R-FCN~\cite{dai2016r}, and Mask R-CNN~\cite{he2017mask}. They operate in two phases, region-proposal and classification~\cite{ren2016faster}. By introducing a Region Proposal Network (RPN) into Faster R-CNN, non-learnable approaches, such as a selective search~\cite{dai2016r}, were removed to further enhance the accuracy of detection, but this came at a cost of greater computational complexity that restricts the use of these methods to a real-time application on UAVs~\cite{he2017mask}. 

To surpass such constraints, the one-stage detectors combine region proposal and classification in the same network to be faster in inference. Representative models include YOLO~\cite{redmon2016you}, Single Shot Multi-Box Detector (SSD)~\cite{liu2016ssd}, and RetinaNet~\cite{uijlings2013selective}. YOLO, introduced by Redmon et al., pioneered real-time detection using a grid-based extractor that predicts bounding boxes and probabilities of each of the grid cells~\cite{redmon2016you}. Subsequent versions, such as YOLOv2, YOLOv3, and YOLOv4, optimized anchor box techniques and backbone models, achieving better accuracy with no loss of speed. SSD and RetinaNet further advanced performance through multi-scale feature maps and better loss functions~\cite{liu2016ssd, lin2017focal}.

Despite these successes, UAV vehicle detection has received difficulties of complex scenarios and detection of small objects. Recent research combines CNNs with architectures such as auto-encoders, and Long Short-Term Memory (LSTM) networks to address these issues. Generative Adversarial Networks (GANs) augment datasets with synthetic images, auto-encoders improve small object detection, and LSTMs capture temporal dependencies in video sequences~\cite{goodfellow2014generative,srivastava2015unsupervised}.

Advanced models like YOLOv3, Faster R-CNN and RetinaNet have won the leader-board in competitions such as VisDrone~\cite{du2019visdrone}. Novel approaches including HRNet~\cite{wang2020deep}, TridentNet~\cite{li2019scale}, and CornerNet~\cite{law2018cornernet} also show promising results, highlighting the ongoing innovation in UAV-based vehicle detection. These developments suggest continued improvements in both detection accuracy and speed as architectures and training methods evolve.

\subsection{Integration with Legacy ITS Frameworks}
A critical barrier to industrial adoption is the interoperability gap between modern UAV perception and legacy Traffic Management Systems such as SCATS or SCOOT. As discussed in architectural frameworks for smart cities~\cite{menouar2017uav}, successful deployment necessitates bridging the disparity between unstructured aerial visual data and the structured inputs required by ground controllers. Future research must therefore prioritize translation middle-ware capable of Virtual Loop Emulation to convert detection coordinates into standard inductive triggers. Furthermore, aligning deep learning outputs with established industry protocols like NTCIP 1202 or DATEX II is essential. While large-scale experiments have validated the utility of massive drone data for traffic modelling~\cite{barmpounakis2020new}, implementing these real-time standards ensures that dynamic aerial insights are directly consumable for signal phase timing, effectively transforming UAVs into plug-and-play sensors.
\section{Future Research Directions}
\label{sec:future_directions}

\subsection{From Detection to Behavioral Analysis} 
Future UAV-ITS frameworks must evolve from simple object detection to predictive behavioral analytics. By integrating machine learning with sophisticated tracking algorithms, systems can move beyond counting vehicles to predicting interactions and potential collisions. For instance, incorporating 3D constrained multiple-kernel tracking~\cite{liu2021deformable} can address frequent occlusions in dense urban canyons. Furthermore, modern systems should incorporate behavioral analysis for proactive traffic management, such as optimizing dynamic traffic signal control based on real-time flow data~\cite{wang2019development}. Modeling multi-vehicle interactions based on speed, direction, and proximity will be pivotal for the next generation of proactive urban traffic management.

\subsection{Multi-UAV Collaboration and Data Fusion}
Deploying multiple UAVs in coordinated systems enhances traffic monitoring through robust collaboration frameworks and advanced data fusion. A central challenge is cross-camera tracking, which requires consistent vehicle IDs across UAV views for accurate trajectory and speed analysis despite variations in angle, scale, lighting, and occlusion. In large-scale scenarios with severe vehicle overlapping, the massive visual data generated by multi-UAV systems can easily saturate traditional communication links. To mitigate these bottlenecks, recent advancements advocate for the integration of advanced communication architectures, such as HAP-enabled edge computing and UAV-mounted Re-configurable Intelligent Surfaces (RIS) within 6G networks. By processing and fusing detection data locally at the edge, these strategies drastically reduce latency and optimize computational load, ensuring resilient and real-time responsiveness in highly dynamic, dense traffic environments~\cite{alotaibi2025optimizing}. As illustrated in Figure~\ref{fig:Cross_cameras_tracking}, vehicles can be re-identified across overlapping fields of view, enabling seamless trajectory reconstruction across multiple UAV perspectives. Du et al. established a benchmark of complex tracking situations with algorithmic testing~\cite{du2018unmanned}, whereas Chen et al. proposed real-time re-identification (Re-ID) models using color, shape, and fine-grained features~\cite{chen2020aware}. Multi-sensor integration supports applications such as taxi demand prediction~\cite{yao2018deep} and UAV swarm-based real-time monitoring with traffic simulations~\cite{zhou2020uav}. Reliable integration of data under different conditions is made possible by effective fusion strategies, like the adaptive fading Unscented Kalman Filter proposed in~\cite{gao2018multi} for multi-sensor state estimation. Multi-UAV systems also have the ability to produce geo-referenced trajectories to enable comprehensive analysis~\cite{sun2018robust, gong2018flight}, and recent developments in large-scale trajectory extraction are achieving new standards in congestion detection, route optimization, and city-wide signal control~\cite{fonod2025advanced}. Cooperative UAV positioning by fusion of multi-sources, as demonstrated by Tang et al., reduces navigation error further~\cite{tang2022multisource}. Ongoing progress in cross-camera tracking, intelligent data fusion, and cooperative positioning is poised to transform real-time traffic management and enhance urban safety. An overview of this process is provided in Figure~\ref{fig:Cross_cameras_tracking}.

\subsection{Small Object Detection}
Small object detection remains a core challenge in UAV-based traffic monitoring due to uneven object scales, sparse spatial distributions, and dense occlusions in aerial imagery~\cite{yang2025lightweight, li2024uav, liao2025atbhc}. Recent advances focus on multidimensional feature extraction, efficient fusion, and lightweight designs to improve accuracy while maintaining low computational cost. LMF-UAV, for example, uses a dual-branch cross- stage universal inverted bottleneck to optimize structural adaptability, achieving a mAP50-95 of 24.6\% on VisDrone with minimal overhead~\cite{yang2025lightweight}. Transformer-based solutions such as UAV-YOLOv5 integrate Swin Transformer V2 modules, Focal-EIOU anchor generation, and BiFormer-based global–local feature capture to boost mAP by 8.5\% over standard YOLOv5~\cite{li2024uav}. Feature enhancement methods like ATBHC-YOLO employ MS-CET for globally sparse feature focus, BHC-FB for scale variance, and WIoU for sample quality assessment, outperforming YOLOv7 by 3.5\%~\cite{liao2025atbhc}. Multi-scale detection improvements, exemplified by YOLOv8-based architectures with weighted down-sampling fusion (WDF) and lightweight heads, deliver up to 43.2\% parameter reduction and 6.4\% mAP50 gains~\cite{sun2024improved}. Ongoing research emphasizes optimizing feature extraction, refining loss functions, and integrating adaptive modules to enhance small-object detection performance while ensuring UAV deployment feasibility.
 
\subsection{Specialized Models for Aerial Data}
Advances in specialized aerial data models are crucial for addressing UAV vehicle detection challenges under varied environmental conditions. Environmental adaptability is particularly central, with a need to ensure accuracy in low light, poor weather, and complex urban environments. Figure~\ref{fig:Illumination} illustrates one such approach, where an illumination-aware classifier selects between specialized detection models to maintain reliable performance across bright, normal, and low-light scenarios. In order to help curb these shortcomings, Gupta et al. examined SLAM-based object detection and scene perception techniques~\cite{gupta2022simultaneous}. Optimization of resources is also a consideration, with model compression and hardware-aware neural architecture search methods decreasing computational requirements while retaining accuracy. Dai et al. introduced the Scale Insensitive Multi-Sensor Fusion Framework (SIMSF) based on graph optimization towards accurate state estimation of the UAVs with a rigid payload constraint~\cite{dai2020simsf}. Sensor fusion serves an additional role of expanding detection accuracy, as indicated by the INS/GPS combination of Nemra and Aouf through State-Dependent Riccati Equation-based filtering~\cite{nemra2010robust} and fusion of inertial measurement data that Noordin et al. rely on to achieve improved stabilization and detection~\cite{noordin2018sensor}. Emphasizing environmental adaptability, efficient computation, and advanced sensor fusion enables the development of robust UAV systems capable of reliable performance in diverse conditions.

\begin{figure}[htbp]
    \centering
    \includegraphics[width=\linewidth]{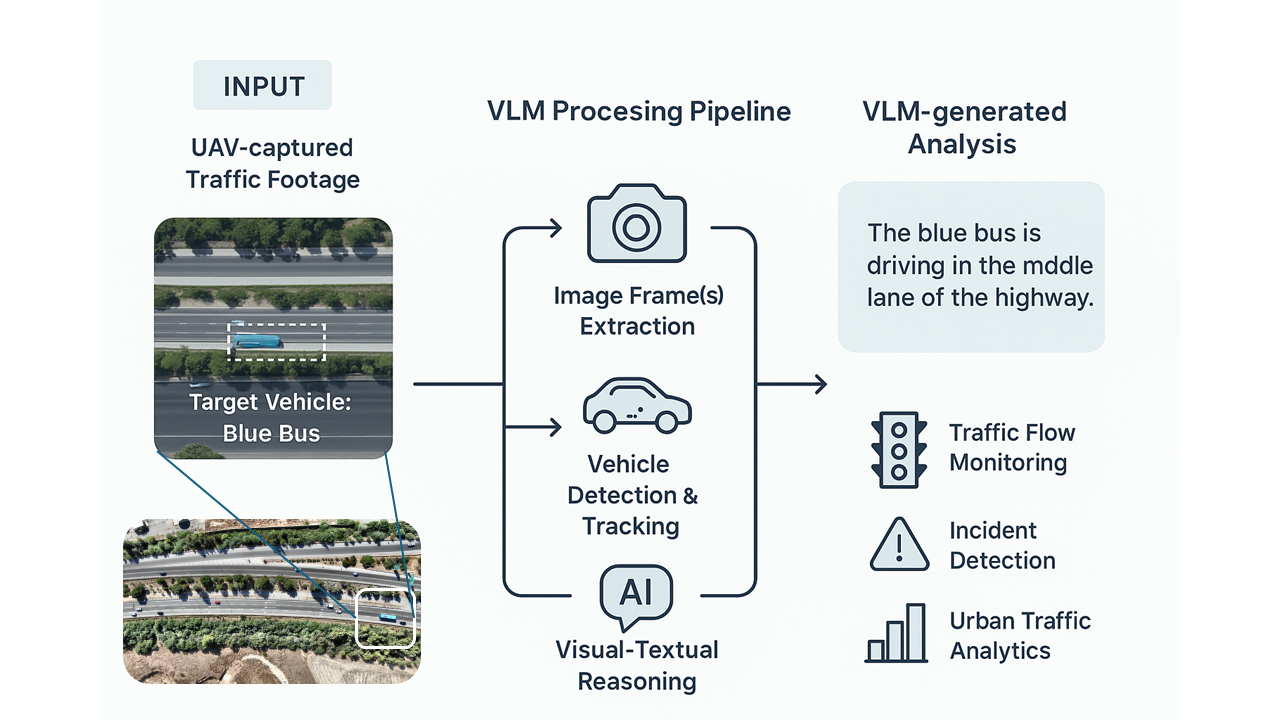}
    \caption{Demonstration of Vision Language Models (VLMs) in traffic monitoring, showcasing the ability to track and describe the movement of a specific vehicle (blue bus) from UAV-recorded highway traffic footage. The VLM generates detailed contextual insights, highlighting its potential for intelligent transport systems.}
    \label{fig:VLM_highway}
\end{figure}

\subsection{Spatio-temporal Video Processing}
Spatio-temporal video processing is essential for UAV-based vehicle detection, especially when combined with ground-based sensors, as it improves both the accuracy and coverage of traffic analysis. He et al. noted that current traffic flow prediction methods struggle to capture the complex spatio-temporal dependencies crucial for ITS~\cite{he2020stnn}. Xu et al. demonstrated that advanced deep learning models significantly improve tracking continuity despite rapid movements and obstructions in UAV car detection~\cite{xu2017car}. Incorporating temporal dynamics into detection architectures has also proven effective: Telegraph et al. proposed a YOLO-based model with temporal augmentation~\cite{telegraph2024spatiotemporal}, and Quan et al. introduced a lightweight multi-frame integration method that preserves real-time performance and improves robustness to motion blur, occlusions, and abrupt appearance changes~\cite{quan2025lightweight}. Tran et al. demonstrated that dynamic resource allocation with 3D Convolutional Networks enhances spatio-temporal feature extraction for action recognition in traffic monitoring~\cite{tran2015learning}. Liu et al. addressed the need to balance computational efficiency and accuracy in video recognition architectures for spatio-temporal applications~\cite{liu2020teinet}. The combination of these advanced approaches with the UAV technology revolutionizes the monitoring and managing of traffic, as well as its real-time responsiveness, safety, and efficiency.

\subsection{Resource-Efficient Computer Vision Optimization for UAV Platforms}
Quantization and optimization of computer vision models are critical for UAV-based intelligent transport systems, addressing constraints in computation, power, and real-time processing. Quantization reduces neural network weight and activation precision to lower computational load and memory usage while preserving detection accuracy~\cite{lee2022yolo}. Zhan et al. presented energy-efficient data collection in the UAV-enabled wireless sensor networks, optimizing the performance and latency subject to the power constraint~\cite{zhan2017energy}. Miniature Machine Learning (MiniML) makes use of compression, lightweight architectures, knowledge distillation, and sparse representation to provide advanced inference on resource-limited platforms. The hardware-aware neural architecture search (NAS) further adapts the models to the UAV hardware and seeks a balance between accuracy, latency, and energy efficiency~\cite{huang2019deep}. Pruning removes unnecessary parameters and adaptive model scaling dynamically scales the model according to the complexity required by the mission. Integrating quantization, MiniML, NAS, and pruning enables UAVs to maintain efficiency during routine surveillance and scale up processing for critical operations, enhancing navigation, traffic monitoring, and emergency response. These combined strategies unlock high-performance, flexible, and energy-efficient aerial vision systems under stringent resource constraints.

\subsection{Simulation-to-Real and Autonomous Scalability} 
Validating UAV advancements in congested urban environments necessitates the use of Digital Twins and high-fidelity simulators (e.g., AirSim, CARLA) to facilitate safe ``Sim-to-Real'' transfer prior to physical deployment. Furthermore, AI integration is pivotal for scaling operations in dynamic settings, driving a paradigm shift from remote-piloted units to fully autonomous swarms. By automating dynamic path planning and collision avoidance in cluttered urban canyons, AI significantly reduces the human-to-UAV ratio, enabling a single operator to oversee large-scale monitoring networks without compromising safety or efficiency.

\subsection{Vision Language Models in Traffic Monitoring}
Vision Language Models (VLMs) are transforming UAV-based traffic monitoring by combining advanced computer vision with natural language processing to enable richer contextual interpretation and semantic reasoning, essential for nuanced analysis of transportation environments. UAVs’ mobility and ability to capture high-resolution aerial imagery make them well-suited for this role, complementing the growing reliance on video-based systems for traffic data collection. As illustrated in Figure~\ref{fig:VLM_highway}, VLMs can simultaneously track vehicles and generate descriptive insights, such as monitoring the movement of a specific bus within highway traffic, thereby extending beyond conventional object detection to provide contextual understanding. For example, the vehicle counting framework proposed by Dai et al. outperforms traditional sensor methods, illustrating VLMs’ potential to improve traffic parameter extraction and monitoring efficiency~\cite{dai2019video}. Future research should integrate advanced VLMs with adaptive learning to address environmental challenges such as variable lighting and occlusions; Theocharides et al. demonstrate real-time traffic state estimation in multi-regional networks, underscoring the importance of contextual information for reliability~\cite{theocharides2024real}. Overall, embedding VLMs in UAV-based systems can enhance analytical capabilities, improve real-time data utilization, and drive more robust, adaptable, and responsive urban traffic surveillance methodologies.

\subsection{Adversarial Attacks and Robustness to Perturbations}
Adversarial attacks pose critical threats to UAV-based computer vision systems for tasks like vehicle detection, exploiting deep learning vulnerabilities through imperceptible perturbations that trigger severe recognition failures and jeopardize safety in applications such as traffic monitoring and autonomous navigation~\cite{yuan2019adversarial}. Neural networks are particularly susceptible, as Goodfellow et al. attribute this to their linear characteristics~\cite{goodfellow2014explaining}, while Papernot et al. highlight black-box attacks that require no model knowledge, intensifying risks in safety-critical UAV deployments~\cite{papernot2017practical}. Empirical studies, including Eykholt et al., show that minor input alterations can mislead recognition systems~\cite{eykholt2018robust}, prompting strategies such as the UAV-specific adversarial recognition methods proposed by Pathak et al.~\cite{pathak2024model}. An explainable deep reinforcement learning approach enables real-time detection of adversarial attacks in UAV guidance and planning with over 90\% accuracy~\cite{hickling2023robust}. Addressing these challenges demands proactive and reactive defenses to strengthen adversarial robustness, as emphasized in broader robustness research~\cite{arnab2018robustness}, ensuring UAV vision systems remain reliable under hostile perturbations.
\section{Conclusion} 
\label{sec:conclusion}
In this study, we presented a comprehensive review of UAV-based vehicle detection techniques, tracing the progression from traditional vision-based methods to state-of-the-art deep learning architectures. We systematically examined how these methods have evolved to meet the unique challenges of aerial traffic monitoring, highlighting the trade-offs between accuracy, computational efficiency, and real-time performance. Particular attention was given to lightweight and medium-scale models, which demonstrate strong potential for UAV deployment under resource-constrained environments. Furthermore, we emphasized the growing integration of detection with vehicle tracking and behavior analysis, a combination that enhances the overall effectiveness of UAV-based traffic monitoring systems. These include the coordination and collaboration of multiple UAVs for large-scale coverage, efficient fusion of multi-sensor data to improve detection reliability, adaptation to complex and dynamic environmental conditions, and fulfilling the stringent requirements of real-time processing. 

Looking ahead, we identified several promising directions for advancing this field. First, spatio-temporal video analysis holds significant potential by leveraging both spatial and temporal cues to improve detection and tracking consistency in dynamic traffic environments. Second, the design of more efficient model architectures, including compact backbones and knowledge distillation strategies, can further reduce computational cost while maintaining competitive accuracy. Third, the development of robust multi-UAV collaboration frameworks, incorporating data from ground-based infrastructure, vehicular sensors, and other UAVs, will play a critical role in enhancing detection reliability and system scalability. Finally, greater attention must be given to system adaptability and resilience, enabling UAV-based solutions to operate effectively in diverse weather, lighting, and urban conditions. By consolidating current knowledge and outlining these future research avenues, this survey provides a foundation for the continued development of robust and scalable UAV-based vehicle detection systems. Ultimately, these advancements will contribute to the realization of intelligent transportation networks that are safer, more efficient, and better equipped to address the complexities of modern urban mobility.

\section*{Acknowledgment}
\label{sec:acknowledgment}
This work was supported by the European Union’s Horizon Europe research and innovation programme under grant agreement No. 101168067 (GuardAI). Views and opinions expressed are however those of the author(s) only and do not necessarily reflect those of the European Union. Neither the European Union nor the granting authority can be held responsible for them.

\bibliographystyle{IEEEtran}
\bibliography{references} 

\vspace{12pt}

\end{document}